%% file: NASOA_with_appendix.tex
\documentclass[10pt,letterpaper,twocolumn]{article}
\usepackage[latin9]{inputenc}
\usepackage{color}
\usepackage{array}
\usepackage{float}
\usepackage{booktabs}
\usepackage{url}
\usepackage{multirow}
\usepackage{amsmath}
\usepackage{amsthm}
\usepackage{amssymb}
\usepackage{graphicx}
\usepackage{wasysym}
\usepackage[unicode=true,
 bookmarks=false,
 breaklinks=true,pdfborder={0 0 1},backref=section,colorlinks=false]
 {hyperref}
\hypersetup{
 pagebackref=true,letterpaper=true,colorlinks}

\makeatletter

\pdfpageheight\paperheight
\pdfpagewidth\paperwidth

\providecommand{\tabularnewline}{\\}
\newcommand{\lyxdot}{.}

\floatstyle{ruled}
\newfloat{algorithm}{tbp}{loa}
\providecommand{\algorithmname}{Algorithm}
\floatname{algorithm}{\protect\algorithmname}

\theoremstyle{plain}
\newtheorem{thm}{\protect\theoremname}
\theoremstyle{remark}
\newtheorem*{rem*}{\protect\remarkname}
\theoremstyle{plain}
\newtheorem*{thm*}{\protect\theoremname}

\usepackage{iccv}
\usepackage{times}
\usepackage{epsfig}
\usepackage{graphicx}

\iccvfinalcopy 

\def\iccvPaperID{2129} 

\ificcvfinal\fi

\usepackage{enumitem}
\usepackage{multicol}

\usepackage{algorithm,algpseudocode}

\usepackage{amsmath,epsfig,amssymb,amsfonts,amsthm,verbatim,epstopdf,float}
\usepackage{url}
\usepackage{bm}
\usepackage{dsfont}
\usepackage{esint}
\usepackage[normalem]{ulem}

\input{math}

\makeatother

\providecommand{\remarkname}{Remark}
\providecommand{\theoremname}{Theorem}

\begin{document}
\title{NASOA: Towards Faster Task-oriented Online Fine-tuning with a Zoo
of Models}
\author{Hang Xu\\
Huawei Noah's Ark Lab\\
\and 
Ning Kang\\
Huawei Noah's Ark Lab\\
\and 
Gengwei Zhang\\
Sun Yat-sen University\\
\and 
Chuanlong Xie\\
Huawei Noah's Ark Lab\\
\and 
Xiaodan Liang$^*$\\
Sun Yat-sen University\\
\and 
Zhenguo Li\\
Huawei Noah's Ark Lab\\
}
\maketitle
\begin{abstract}
\vspace{-2mm}
\let\thefootnote\relax\footnotetext{$^*$ Corresponding Author: xdliang328@gmail.com}Fine-tuning
from pre-trained ImageNet models has been a simple, effective, and
popular approach for various computer vision tasks. The common practice
of fine-tuning is to adopt a default hyperparameter setting with a
fixed pre-trained model, while both of them are not optimized for
specific tasks and time constraints. Moreover, in cloud computing
or GPU clusters where the tasks arrive sequentially in a stream, faster
online fine-tuning is a more desired and realistic strategy for saving
money, energy consumption, and CO2 emission. In this paper, we propose
a joint Neural Architecture Search and Online Adaption framework named
NASOA towards a faster task-oriented fine-tuning upon the request
of users. Specifically, NASOA first adopts an offline NAS to identify
a group of training-efficient networks to form a pretrained model
zoo. We propose a novel joint block and macro level search space to
enable a flexible and efficient search. Then, by estimating fine-tuning
performance via an adaptive model by accumulating experience from
the past tasks, an online schedule generator is proposed to pick up
the most suitable model and generate a personalized training regime
with respect to each desired task in a one-shot fashion.  The resulting
model zoo$^{1}$\footnote{$^{1}$The efficient training model zoo (ET-NAS) has been released
at: \url{https://github.com/NAS-OA/NASOA}} is more training efficient than SOTA models, e.g. 6x faster than
RegNetY-16GF, and 1.7x faster than EfficientNetB3. Experiments on
multiple datasets also show that NASOA achieves much better fine-tuning
results, i.e. improving around 2.1\% accuracy than the best performance
in RegNet series under various constraints and tasks; 40x faster compared
to the BOHB.
\end{abstract}

\section{\vspace{-2mm}
Introduction}

Fine-tuning using pre-trained models becomes the de-facto standard
in the field of computer vision because of its impressive results
on various downstream tasks such as fine-grained image classification
\cite{nilsback2008automated,WelinderEtal2010}, object detection \cite{He2018,jiang2018hybrid,Xu_2019_ICCV}
and segmentation \cite{chen2017deeplab,liu2019auto}. \cite{kornblith2019better,He2018}
verified that fine-tuning pre-trained networks outperform training
from scratch. It can further help to avoid over-fitting \cite{cui2018large}
as well as reduce training time significantly \cite{He2018}. Due
to those merits, many cloud computing and AutoML pipelines provide
fine-tuning services for an online stream of upcoming users with new
data, different tasks and time limits. In order to save the user's
time, money, energy consumption, or even CO2 emission, an efficient
online automated fine-tuning framework is practically useful and in
great demand. Thus, we propose to explore faster online fine-tuning.

The conventional practice of fine-tuning is to adopt a set of predefined
hyperparameters for training a predefined model \cite{li2020rethinking}.
It has three drawbacks in the current online setting: 1) The design
of the backbone model is not optimized for the upcoming fine-tuning
task and the selection of the backbone model is not data-specific.
2) A default setting of hyperparameters may not be optimal across
tasks and the training settings may not meet the time constraints
provided by users. 3) With the incoming tasks, the regular diagram
is not suitable for this online setting since it cannot memorize and
accumulate experience from the past fine-tuning tasks. Thus, we propose
to decouple our faster fine-tuning problem into two parts: finding
efficient fine-tuning networks and generating optimal fine-tuning
schedules pertinent to specific time constraints in an online learning
fashion.

Recently, Neural Architecture Search (NAS) algorithms demonstrate
promising results on discovering top-accuracy architectures, which
surpass the performance of hand-crafted networks and saves human's
efforts \cite{zoph2018learning,liu2018progressive,liu2018darts,radosavovic2019network,tan2019efficientdet,Real2019,Tan2019,yao2019sm}
as well as studying NAS across tasks and datasets \cite{chen2020catch,duan2021transnas}.
However, those NAS works usually focus on inference time/FLOPS optimization
and their search space is not flexible enough which cannot guarantee
the optimality for fast fine-tuning. In contrast, we resort to developing
a NAS scheme with a novel flexible search space for fast fine-tuning.
On the other hand, hyperparameter optimization (HPO) methods such
as grid search \cite{bergstra2012random}, Bayesian optimization (BO)
\cite{snoek2012practical,mendoza2016towards}, and BOHB \cite{falkner2018bohb}
are used in deep learning and achieve good performance. However, those
search-based methods are computationally expensive and require iterative
``trial and error'', which violate our goal for faster adaptation
time. 

In this work, we propose a novel Neural Architecture Search and Online
Adaption framework named NASOA. First, we conduct an offline NAS for
generating an efficient fine-tuning model zoo. We design a novel\textbf{
}block-level and macro-structure search space\textbf{ }to allow a
flexible choice of the networks. Once the efficient training model
zoo is created offline NAS by Pareto optimal models, the online user
can enjoy the benefit of those efficient training networks without
any marginal cost. We then propose an online learning algorithm with
an adaptive predictor to modeling the relation between different hyperparameter,
model, dataset meta-info and the final fine-tuning performance. The
final training schedule is generated directly from selecting the fine-tuning
regime with the best predicted performance. Benefiting from the experience
accumulation via online learning, the diversity of the data and the
increasing results can further continuously improve our regime generator.
Our method behaves in a one-shot fashion and doesn't involve additional
searching cost as HPO, endowing the capability of providing various
training regimes under different time constraints. We also theoretically
prove the convergence of the optimality of our proposed online model.

Extensive experiments are conducted on multiple widely used fine-tuning
datasets. The searched model zoo ET-NAS is more training efficient
than SOTA ImageNet models, e.g. 5x training faster than RegNetY-16GF,
and 1.7x faster than EfficientNetB3. Moreover, by using the whole
NASOA, our online algorithm achieves superior fine-tuning results
in terms of both accuracy and fine-tuning speed, i.e. improving around
2.1\% accuracy than the best performance in RegNet series under various
tasks; saving 40x computational cost comparing to the BOHB method.

Our contributions are summarized as follows:\vspace{-1mm}

\begin{itemize}[leftmargin=*]

\item We make the first effort to propose a faster fine-tuning pipeline
that seamlessly combines the training-efficient NAS and online adaption
algorithm. Our NASOA can effectively generate a personalized fine-tuning
schedule of each desired task via an adaptive model for accumulating
experience from the past tasks.\vspace{-1mm}

\item The proposed novel joint block/macro level search space enables
a flexible and efficient search. The resulting model zoo ET-NAS is
more training efficient than very strong ImageNet SOTA models e.g.
EfficientNet, RegNet. All the ET-NAS models have been released to
help the community skipping the computation-heavy NAS stage and directly
enjoy the benefit of NASOA.\vspace{-1mm}

\item The whole NASOA pipeline achieves much better fine-tuning results
in terms of both accuracy and fine-tuning efficiency than current
fine-tuning best practice and HPO method, e.g. , 40x faster compared
to the BOHB method.

\end{itemize}

\section{Related Work\vspace{-2mm}
}

\textbf{Neural Architecture Search (NAS).} The goal of NAS is to automatically
optimize network architecture and release human effort from this handcraft
network architecture engineering. Most previous works \cite{liu2018darts,cai2018proxylessnas,liu2018progressive,tan2018mnasnet,xie2018snas,howard2019searching}
aim at searching for CNN architectures with better inference and fewer
FLOPS. \cite{baker2016designing,cai2018efficient,zhong2018practical}
apply reinforcement learning to train an RNN controller to generate
a cell architecture. \cite{liu2018darts,xie2018snas,cai2018proxylessnas}
try to search a cell structure by weight-sharing and differentiable
optimization. \cite{Tan2019} use a grid search for an efficient network
by altering the depth/width of the network with a fixed block structure.
On the contrary, our NAS focuses creating an efficient training model
zoo for fast fine-tuning. Moreover, the existing search space design
cannot meet the purpose of our search.

\textbf{Generating Hyperparameters for Fine-tuning.} HPO methods such
as Bayesian optimization (BO) \cite{snoek2012practical,mendoza2016towards},
BOHB \cite{falkner2018bohb} achieves very promising result but require
a lot of computational resources which is contradictory to our original
objective of efficient fine-tuning. On the other hand, limited works
discuss the model selection and HPO for fine-tuning. \cite{kornblith2019better}
finds that ImageNet accuracy and fine-tuning accuracy of different
models are highly correlated. \cite{li2020rethinking,achille2019task2vec}
suggest that the optimal hyperparameters and model for fine-tuning
should be both dataset dependent and domain similarity dependent \cite{cui2018large}.
HyperStar \cite{mittal2020hyperstar} is a concurrent HPO work demonstrating
that a performance predictor can effectively generate good hyper-parameters
for a single model. However, those works don't give an explicit solution
about how to perform fine-tuning in a more practical online scenario.
In this work, we take the advantage of online learning \cite{hoi2018online,sahoo2017online}
to build a schedule generator, which allows us to memorize the past
training history and provide up-and-coming training regimes for new
coming tasks on the fly. Besides, we introduce the NAS model zoo to
further push up the speed and performance.

\begin{figure*}
\begin{centering}
\vspace{-1mm}
\includegraphics[width=2\columnwidth]{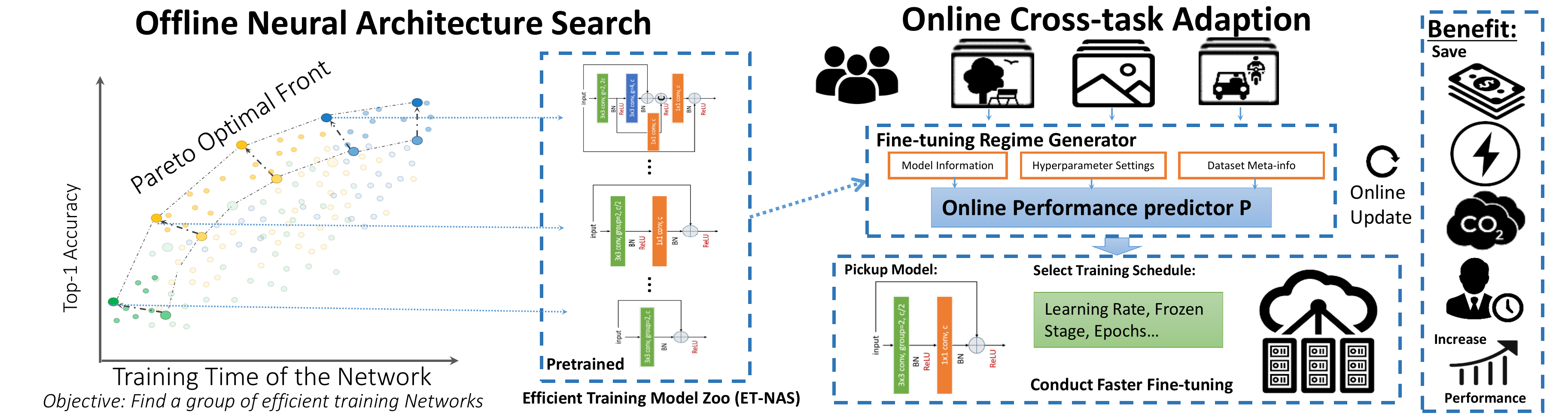}
\par\end{centering}
\vspace{-1mm}
\caption{\label{fig:Overview}Overview of our NASOA. Our faster task-oriented
online fine-tuning system has two parts: a) Offline NAS to generate
an efficient training model zoo with good accuracy and training speed;
b) An online fine-tuning regime generator to perform a task-specific
fine-tuning with a suitable model under user's time constraint.}

\centering{}\vspace{-1mm}
\end{figure*}

\section{The Proposed Approach\vspace{-2mm}
}

The goal of this paper is to develop an online fine-tuning pipeline
to facilitate a fast continuous cross-task model adaption. By the
preliminary experiments in Section \ref{subsec:Preliminary-Experiments},
we confirm that the model architectures and hyperparameters such as
the learning rate and frozen stages will greatly influence the accuracy
and speed of the fine-tuning program. Thus, our NASOA includes two
parts as shown in the Figure \ref{fig:Overview}: 1) Searching a group
of neural architectures with good accuracy and fast training speed
to create a pretrained model zoo; 2) Designing an online task-oriented
algorithm to generate an efficient fine-tuning regime with the most
suitable model under user's time constraint.

\subsection{Creating an Efficient Training Model Zoo (ET-NAS) by NAS \label{subsec:Component-1:-Create_model_zoo}}

The commonly used hand-craft backbones for fine-tuning including MobileNet
\cite{sandler2018mobilenetv2}, ResNet \cite{he2016deep}, and ResNeXt
\cite{xie2017aggregated}. Recently, some state-of-the-art backbone
series such as RegNet \cite{radosavovic2020designing}, and EfficientNet
\cite{tan2019efficientdet} are developed by automated algorithms
for higher accuracy and faster inference speed. However, the objective
of our NAS is to find a group of models with good model generalization
ability and training speed. Suggested by \cite{kornblith2019better},
the model fine-tuning accuracy (model generalization ability) has
a strong correlation between ImageNet accuracy ($r=0.96$). Meanwhile,
the training speed can be measured by the step time of each training
iteration. Thus, our NAS can be formulated by a multi-objective optimization
problem (MOOP) on the search space $S$ given by:{\small{}
\begin{equation}
\max_{\mathcal{A}\in S}\text{\ensuremath{\left(\mathrm{acc}(\mathcal{A}),-T_{s}(\mathcal{A})\right)} subject to }\:T_{s}(\mathcal{A})\leq T_{m}\label{eq:MOOP}
\end{equation}
}where $\mathcal{A}$ is the architecture, $\mathrm{acc}(.)$ is the
Top-1 accuracy on ImageNet, $T_{s}(.)$ is the average step time of
one iteration, and $T_{m}$ is the maximum step time allowed. The
step time is defined to be the total time of one iteration, including
forward/backward propagation, and parameter update.

\begin{figure*}
\begin{centering}
\vspace{-13mm}
\includegraphics[width=1.8\columnwidth]{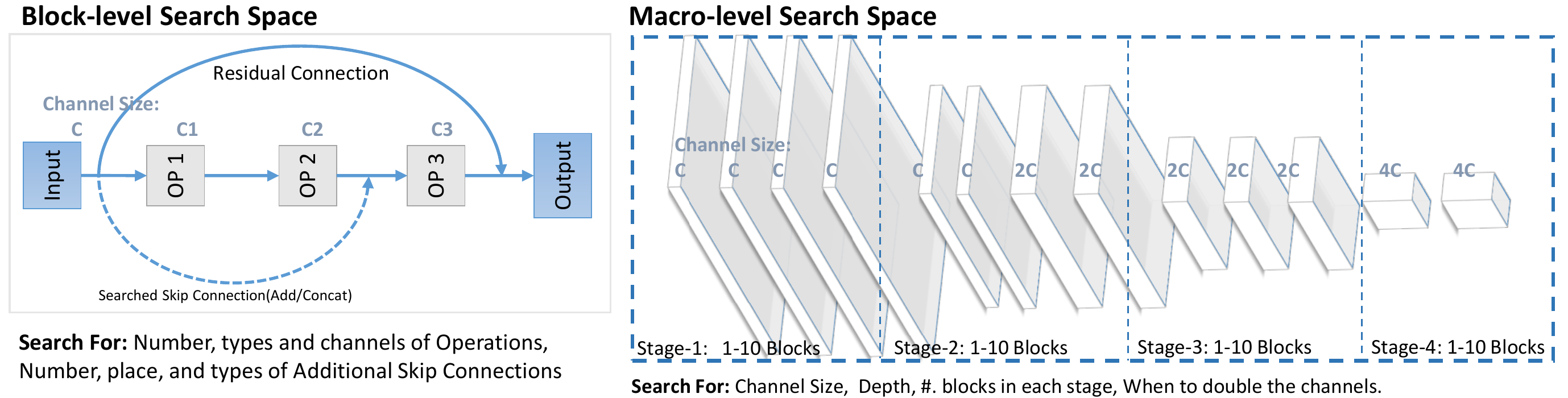}
\par\end{centering}
\vspace{-1mm}
\caption{\label{fig:Search_space}Our joint block/macro-level search space
to find efficient training networks. Our block-level search space
covers many popular designs such as ResNet, ResNext, MobileNet Block.
Our macro-level search space allows small adjustment of the network
in each stage thus the resulting models are more flexible and efficient.}

\centering{}\vspace{-1mm}
\end{figure*}

\textbf{Search Space Design} is extremely important \cite{radosavovic2020designing}.
As shown in Figure \ref{fig:Search_space}, we propose a novel flexible
joint block-level and macro-level search space to enable simple to
complex block design and fine adjustment of the computation allocation
on each stage. Unlike existing topological cell-level search space
such as DARTS \cite{liu2018darts}, AmoebaNet\cite{Real2019}, and
NASBench101\cite{dong2019bench}, ours is more compact and avoids
redundant skip-connections which have great memory access cost (MAC).
Our block-level search space is more flexible to adjust the width,
depth (for each stage), when to down-sample/raise the channels. In
contrast, EfficientNet only scales up/down the total width and depth
by a fixed allocation ratio, and RegNet cannot change the number/type
of operations in each block. 

\textbf{Block-level Search Space. }We consider a search space based
on 1-3 successive nodes of 5 different operations. Three skip connections
with one fixed residual connection are searched. Element-wise add
or channel-wise concat is chosen to combine the features for the skip-connections.
For each selected operation, we also search for the ratio of changing
channel size: $\times0.25$, $\times0.5$, $\times1$, $\times2$,
$\times4$. Note that it can cover many popular block designs such
as Bottleneck \cite{he2016deep}, ResNeXt \cite{xie2017aggregated}
and MB block \cite{sandler2018mobilenetv2}. It consists of $5.4\times10^{6}$
unique blocks. 

\textbf{Macro-level Search Space. }Allocation computation over different
stages is crucial for a backbone \cite{liang2019computation}. Early-stage
feature maps in one backbone are larger which captures texture details,
while late-stage feature maps are smaller which are more discriminative
\cite{Li2018}. Therefore, for macro-level search space, we design
a flexible search space to find the optimal channel size (width),
depth (total number of blocks), when to down-sample, and when to raise
the channels. Our macro-level structure consists of 4 flexible stages.
The spatial size of the stages is gradually down-sampled with factor
2. In each stage, we stack a number of block architectures. The positions
of the doubling channel block are also flexible. This search space
consists of $1.5\times10^{7}$ unique architectures. 

\textbf{Multi-objective Searching Algorithm}. For MOOP in Eq \ref{eq:MOOP},
we define architecture $\mathcal{A}_{1}$ \textit{dominates} $\mathcal{A}_{2}$
if (i) $\mathcal{A}_{1}$ is no worse than $\mathcal{A}_{2}$ in all
objectives; (ii) $\mathcal{A}_{1}$ is strictly better than $\mathcal{A}_{2}$
in at least one objective. $\mathcal{A}^{*}$ is \textit{Pareto optimal}
if there is no other $\mathcal{A}$ that dominate $\mathcal{A}^{*}$.
The set of all \textit{Pareto optimal} architectures constitutes the
\textit{Pareto front}. To solve this MOOP problem, we modify a well-known
method named Elitist Non-Dominated sorting genetic algorithm (NSGA-II)
\cite{deb2000fast} to optimize the \textit{Pareto front} $\mathcal{P}_{f}$
.  The main idea of NSGA-II is to rank the sampled architectures
by non-dominated sorting and preserve a group of elite architectures.
Then a group of new architectures is sampled and trained by mutation
of the current elite architectures on the $\mathcal{P}_{f}$ . The
algorithm can be paralleled on multiple computation nodes and lift
the $\mathcal{P}_{f}$ simultaneously. We modify the NSGA-II algorithm
to become a NAS algorithm: a) To enable parallel searching on N computational
nodes, we modify the non-dominated-sort method to generate exactly
N mutated models for each generation, instead of a variable size as
the original NSGA-II does. b) We define a group of mutation operations
for our block/macro search space for NSGA-II to change the network
structure dynamically. c) We add a parent computation node to measure
the selected architecture's training speed and generate the $\mathcal{P}_{f}$
. 

\textbf{Efficient Training Model Zoo $Z_{oo}$ (ET-NAS). }By the proposed
NAS method, we then create an efficient-training model zoo $Z_{oo}$
named \textbf{ET-NAS} which consists of $K$ \textit{Pareto optimal}
models $\mathcal{A}_{i}^{*}$ on $\mathcal{P}_{f}$. $\mathcal{A}_{i}^{*}$
are pretrained by ImageNet. Details of the search space, its encoding,
NSGA-II algorithm and $\mathcal{A}_{i}^{*}$ architectures can be
found in Appendix.

\subsection{\vspace{-1mm}
Online Task-oriented Fine-tuning Schedule Generation\label{sec:Component-2:-Generating}}

With the help of efficient training $Z_{oo}$, the marginal computational
cost of each user is minimized while they can enjoy the benefit of
NAS. We then need to decide a suitable fine-tuning schedule upon the
user\textquoteright s upcoming tasks. Given user's dataset $D$ and
fine-tuning\textbf{ }time constraint $T_{l}$, an online regime generator
$G(.,.)$ is desired: 
\begin{align}
\left[\mathrm{Regime_{FT}},\mathcal{A}_{i}^{*}\right] & =G(D,T_{l}),\label{eq:regimegenearator}
\end{align}
\vspace{-2mm}
\[
\textrm{ such that }\mathrm{Acc}(\mathcal{A}_{i}^{FineTune},D_{val})\textrm{ is maximized,}
\]
where the $\mathrm{Regime_{FT}}$ includes all the hyperparameters
required, i.e., $lr$ schedule, total training steps, and frozen stages.
$G(.,.)$ also needs to pick up the most suitable pretrained model
$\mathcal{A}_{i}^{*}$ from $Z_{oo}$. Note that existing search-based
HPO methods require huge computational resources and cannot fit in
our online one-shot training scenario. Instead, we first propose an
online learning predictor $Acc_{P}$ to model the accuracy on the
validation set $\mathrm{Acc}(\mathcal{A}_{i}^{FT},D_{val})$ by the
meta-data information. Then we can use the predictor to construct
$G(.,.)$ to generate an optimal hyperparameter setting and model.

\subsubsection{Online Learning for Modeling $\mathrm{Acc}(\mathcal{A}_{i}^{FT},D_{val})$\label{subsec:Online-Learning-for}}

Recently, \cite{li2020rethinking} suggest that the optimal hyperparameters
for fine-tuning are highly related to some data statistics such as
domain similarity to the ImageNet. Thus, we hypothesis that we can
model the final accuracy by a group of predictors, e.g., model information,
meta-data description, data statistics $\mathrm{stat}(D)$, domain
similarity, and hyperparameters. We list the variables we considered
to predict the accuracy result as follows:
\begin{center}
\vspace{-2mm}
\renewcommand\arraystretch{0.9}\tabcolsep 0.2in{\scriptsize{}}%
\begin{tabular}{|cc|}
\hline 
\multicolumn{2}{|c|}{{\scriptsize{}Model $\mathcal{A}_{i}^{*}$ name (one-hot dummy variable)}}\tabularnewline
\multicolumn{2}{|c|}{{\scriptsize{}Domain Similiarity to ImageNet (EMD) \cite{cui2018large}}}\tabularnewline
{\scriptsize{}Average \#. images per class} & {\scriptsize{}Std \#. images per class}\tabularnewline
\multicolumn{2}{|c|}{{\scriptsize{}ImageNet Acc. of the $\mathcal{A}_{i}^{*}$}}\tabularnewline
{\scriptsize{}\#.Classes} & {\scriptsize{}Number of Iteration}\tabularnewline
{\scriptsize{}Learning Rate} & {\scriptsize{}Frozen Stages}\tabularnewline
\hline 
\end{tabular}\vspace{-1mm}
\par\end{center}

Those variables can be easily calculated ahead of the fine-tuning.
One can prepare offline training data by fine-tuning different kinds
of dataset and collect the accuracy correspondingly and apply a Multi-layer
Perceptron Regression (MLP) offline on it. However, online learning
should be a more realistic setting for our problem. In cloud computing
service or a GPU cluster, a sequence of fine-tuning requests with
different data will arrive from time to time. The predictive model
can be further improved by increasing the diversity of the data and
the requests over time.

Using a fixed depth of MLP model in the online setting may be problematic.
Shallow networks maybe more preferred for small number of instances,
while deeper model can achieve better performance when the sample
size becomes larger. Inspired by \cite{sahoo2017online}, we use an
adaptive MLP regression to automatically adapt its model capacity
from simple to complex over time. Given the input variables, the prediction
of the accuracy is given by:{\small{}
\begin{align}
Acc_{P}(\mathcal{A}_{i}^{*},\mathrm{Regime_{FT}},stat(D)) & =\sum_{l=1}^{L}\alpha_{l}\mathrm{f}_{l},\label{eq:adaptive_MLP}
\end{align}
}\vspace{-1mm}
{\footnotesize{}
\[
\mathrm{f}_{l}=h_{l}W_{l},\;h_{l}=RELU(\varPhi_{l}h_{l-1}),\;h_{0}=[\mathcal{A}_{i}^{*},\mathrm{Regime_{FT}},\mathrm{stat}(D)].
\]
}where $\textrm{ }l=0,...,L$. The predicted accuracy is a weighted
sum of the output $\mathrm{f}_{l}$ of each intermediate fully-connected
layer $h_{l}$. The $W_{l}$ and $\varPhi_{l}$ are the learnable
weights of each fully-connected layer. The $\alpha_{l}$ is a weight
vector assigning the importance to each layer and $\left\Vert \alpha\right\Vert =1$.
Thus the predictor $Acc_{P}$ can automatically adapt its model capacity
from simple to complex along with incoming tasks. The learnable weight
$\alpha_{l}$ controls the importance of each intermediate layer and
the final predicted accuracy is a weighted sum of $\mathrm{f}_{l}$
of them. The network can be updated by a Hedge Backpropagation \cite{freund1999adaptive}
in which $\alpha_{l}$ is updated based on the loss suffered by this
layer $l$ as follows:{\small{}
\begin{equation}
\alpha_{l}'\leftarrow\alpha_{l}\beta^{\mathcal{L}(\mathrm{f}_{l},Acc_{gt})},\:W_{l}'\leftarrow W_{l}-\eta\alpha_{l}\nabla_{W_{l}}\mathcal{L}(\mathrm{f}_{l},Acc_{gt})\label{eq:back-prop}
\end{equation}
}\vspace{-1mm}
{\small{}
\[
\varPhi_{l}'\leftarrow\varPhi_{l}'-\eta\sum_{j=l}^{L}\alpha_{j}\nabla_{W_{l}}\mathcal{L}(\mathrm{f}_{j},Acc_{gt}),\:\alpha_{l}''\leftarrow\frac{\alpha_{l}'}{\sum\alpha_{l}'}
\]
}where $\beta\in(0,1)$ is the discount rate, the weight $\alpha_{l}'$
are re-normalized such that $\left\Vert \alpha\right\Vert =1$, and
$\eta$ is the learning rate. Thus, during the online update, the
model can choose an appropriate depth by $\alpha_{l}$ based on the
performance of each output at that depth. By utilizing the online
cumulative results, our generator gains experience that helps future
prediction. 

\textbf{Generating Task-oriented Fine-tuning Schedule.} Our schedule
generator $G$ then can make use of the performance predictor to find
the best training regime $G(D,T_{l})$: {\footnotesize{}
\[
\mathrm{\arg\max}_{\mathcal{A}\in Z_{oo},Regime_{FT}\in S_{FT}}Acc_{P}(\mathcal{A},Regime_{FT},\mathrm{stat}(D)).
\]
}Once the time constraint $T_{l}$ is provided, the max number of
iterations for different $\mathcal{A}_{i}^{*}$ can be calculated
by an offline step-time lookup table for $Z_{oo}$. The corresponding
meta-data variables can be then calculated for the incoming task.
The optimal selection of model and hyperparameters is obtained by
ranking the predicted accuracy of all possible grid combinations.
Details can be found in the Appendix.

\textbf{Theoretical Analysis. }Let the $\alpha$ and $\mathcal{L}$
denote as{\small{}
\[
\mathcal{\alpha}=\left(\alpha_{1},\alpha_{2},\ldots,\alpha_{L}\right)^{\mathrm{T}},\mathcal{L}=\left(\mathcal{L}_{1},\mathcal{L}_{2},\ldots,\mathcal{L}_{L}\right)^{\mathrm{T}}
\]
} where $\mathcal{L}_{l}=\mathcal{L}(f_{l},Acc_{gt})$ for $l=1,2,\ldots,L.$
At time $0\leq t\leq T$, we denote $\alpha$ and $\mathcal{L}$ as
$\alpha^{(t)}$ and $\mathcal{L}^{(t)}$ respectively.
\begin{thm}
Suppose the number of layers $L$ is a fixed integer, the training
time $T$ is sufficiently large and the loss function $\mathcal{L}(f_{l},Acc_{gt})$
is bounded in $[0,1]$. The sequence of the weight vectors: $\{\alpha^{(1)},\alpha^{(1)},\ldots,\alpha^{(T)}\},$
is learned by the Hedge Backpropagation in (\ref{eq:back-prop}).
The initialized weight vector $\alpha^{(1)}$ is the uniform discrete
distribution $\alpha^{(1)}=(\frac{1}{L},\frac{1}{L},\ldots,\frac{1}{L}).$
The discount rate $\beta$ is fixed during the training procedure
and is taken to be $\sqrt{T}/(\sqrt{T}+C)$ given $T$, where $C$
is a fixed constant. Then the average regret of the online learning
algorithm for modelling $Acc(\mathcal{A}_{i}^{FT},D_{val})$ satisfies
\[
\frac{1}{T}\sum_{t=1}^{T}\left(\alpha^{(t)}\right)^{\mathrm{T}}\mathcal{L}^{(t)}-\min_{\alpha}\frac{1}{T}\sum_{t=1}^{T}\alpha^{\mathrm{T}}\mathcal{L}^{(t)}\leq O(\frac{1}{\sqrt{T}}).
\]
\end{thm}

\begin{proof}
The detailed proof is in the Appendix.
\end{proof}
\begin{rem*}
This theorem shows that the empirical average regret of the learned
sequence $\{\alpha^{(t)},t=1,\ldots,T\}$ converges to the optimal
achievable average regret as $T$ tends to infinity. For any given
$\alpha'$ such that
\[
\sum_{t=1}^{T}\left(\alpha'\right)^{\mathrm{T}}\mathcal{L}^{(t)}-\min_{\alpha}\sum_{t=1}^{T}\alpha^{\mathrm{T}}\mathcal{L}^{(t)}>0
\]
 the learnt weight vector $\alpha^{(t)}$ finally outperforms $\alpha'$
as long as the training time $T$ is sufficiently large. Obviously,
$\alpha'$ can be any one-hot vector. This implies our adaptive learning
is \textbf{better than the regression with a fixed-depth neural network}.
Hence, after accumulating enough experience, \textbf{the online learning
procedure finds out a solution that approaches the optimality}. The
learned weight vector $\alpha^{(t)}$ can capture the optimal model
capacity that fully employs the the power of depth to learn complex
patterns and also guarantees the faster convergence rate of a shallow
model.
\end{rem*}

\section{\vspace{-1mm}
Experimental Results\vspace{-1mm}
}

\subsection{Preliminary Experiments\label{subsec:Preliminary-Experiments}}

\begin{table*}
\begin{centering}
\vspace{-2mm}
\par\end{centering}
\begin{centering}
\renewcommand\arraystretch{1}\tabcolsep 0.06in{\footnotesize{}}%
\begin{tabular}{ccccc|ccccc}
\hline 
{\footnotesize{}DataSets} & {\footnotesize{}\#.Class} & {\footnotesize{}Task} & {\footnotesize{}\#.Train} & {\footnotesize{}\#.Test} & {\footnotesize{}DataSets} & {\footnotesize{}\#.Class} & {\footnotesize{}Task} & {\footnotesize{}\#.Train} & {\footnotesize{}\#.Test}\tabularnewline
\hline 
\textbf{\footnotesize{}Flowers102}{\footnotesize{}\cite{nilsback2008automated}} & {\footnotesize{}102} & {\footnotesize{}Fine-Grained} & {\footnotesize{}6K} & {\footnotesize{}2K} & {\footnotesize{}Stanford-Car\cite{KrauseStarkDengFei-Fei_3DRR2013}} & {\footnotesize{}196} & {\footnotesize{}Fine-Grained} & {\footnotesize{}8K} & {\footnotesize{}8K}\tabularnewline
\textbf{\footnotesize{}CUB-Birds}{\footnotesize{}\cite{WelinderEtal2010}} & {\footnotesize{}200} & {\footnotesize{}Fine-Grained} & {\footnotesize{}10K} & {\footnotesize{}2K} & {\footnotesize{}MIT67\cite{quattoni2009recognizing}} & {\footnotesize{}67} & {\footnotesize{}Scene cls.} & {\footnotesize{}5K} & {\footnotesize{}1K}\tabularnewline
\textbf{\footnotesize{}Caltech101}{\footnotesize{}\cite{fei2006one}} & {\footnotesize{}101} & {\footnotesize{}General} & {\footnotesize{}8K} & {\footnotesize{}1K} & {\footnotesize{}Food101\cite{bossard2014food}} & {\footnotesize{}101} & {\footnotesize{}Fine-Grained} & {\footnotesize{}75K} & {\footnotesize{}25K}\tabularnewline
\textbf{\footnotesize{}Caltech256}{\footnotesize{}\cite{griffin2007caltech}} & {\footnotesize{}257} & {\footnotesize{}General} & {\footnotesize{}25K} & {\footnotesize{}6K} & {\footnotesize{}FGVC Aircrafts\cite{maji2013fine}} & {\footnotesize{}100} & {\footnotesize{}Fine-Grained} & {\footnotesize{}7K} & {\footnotesize{}3K}\tabularnewline
\textbf{\footnotesize{}Stanford-Dog}{\footnotesize{}\cite{KhoslaYaoJayadevaprakashFeiFei_FGVC2011}} & {\footnotesize{}120} & {\footnotesize{}Fine-Grained} & {\footnotesize{}12K} & {\footnotesize{}8K} & {\footnotesize{}Blood-cell\cite{singh2020blood}} & {\footnotesize{}4} & {\footnotesize{}Medical Img.} & {\footnotesize{}10K} & {\footnotesize{}2K}\tabularnewline
\hline 
\end{tabular}{\footnotesize\par}
\par\end{centering}
\begin{centering}
\caption{\label{tab:Datasets-and-their}Datasets and their statistics used
in this paper. Datasets in bold are used to construct the online learning
training set. The rest are used to test our NASOA. It is commonly
believed that Aircrafts, Flowers102 and Blood-cell deviate from the
ImageNet domain.}
\par\end{centering}
\centering{}\vspace{-1mm}
\end{table*}

We conduct a complete preliminary experiment to justify our motivation
and model settings. Details can be found in the Appendix. According
to our experiments, we find that for an efficient fine-tuning, the
model matters most. \textbf{The suitable model should be selected
according to the task and time constraints}. Thus constructing a model
zoo with various sizes of training-efficient models and picking up
suitable models should be a good solution for faster fine-tuning.
We also verify some existing conclusions: Fine-tuning performs better
than training from scratch \cite{kornblith2019better} so that our
topic is very important for efficient GPU training; Learning rate
and frozen stage are crucial for fine-tuning \cite{guo2019spottune},
which needs careful adjustment. 

\subsection{\vspace{-1mm}
Offline NAS and Model Zoo Results\label{subsec:NAS-Implementation-Details} }

\begin{table}
\begin{centering}
\vspace{-2mm}
\par\end{centering}
\begin{centering}
\renewcommand\arraystretch{1}\tabcolsep 0.06in{\scriptsize{}}%
\begin{tabular}{ccccc}
\hline 
\multirow{2}{*}{{\scriptsize{}Model Name}} & {\scriptsize{}Top-1} & {\scriptsize{}Inf} & {\scriptsize{}Training Step} & {\scriptsize{}Training GPU}\tabularnewline
 & {\scriptsize{}Acc.} & {\scriptsize{}Time (ms)} & {\scriptsize{}Time (ms)} & {\scriptsize{}Usage (MB)}\tabularnewline
\hline 
{\scriptsize{}RegNetY-200MF\cite{radosavovic2020designing}} & {\scriptsize{}70.40} & {\scriptsize{}14.25} & {\scriptsize{}62.30} & {\scriptsize{}2842}\tabularnewline
\textbf{\scriptsize{}ET-NAS-C} & \textbf{\scriptsize{}71.29} & \textbf{\scriptsize{}8.94} & \textbf{\scriptsize{}26.28} & \textbf{\scriptsize{}2572}\tabularnewline
\hline 
{\scriptsize{}RegNetY-400MF\cite{radosavovic2020designing}} & {\scriptsize{}74.10} & {\scriptsize{}20.57} & {\scriptsize{}90.61} & {\scriptsize{}4222}\tabularnewline
\textbf{\scriptsize{}ET-NAS-D} & \textbf{\scriptsize{}74.46} & \textbf{\scriptsize{}14.54} & \textbf{\scriptsize{}36.30} & \textbf{\scriptsize{}3184}\tabularnewline
\hline 
{\scriptsize{}RegNetY-600MF\cite{radosavovic2020designing}} & {\scriptsize{}75.50} & {\scriptsize{}22.15} & {\scriptsize{}90.11} & \textbf{\scriptsize{}4498}\tabularnewline
{\scriptsize{}MobileNet-V3-Large\cite{howard2019searching}} & {\scriptsize{}75.20} & \textbf{\scriptsize{}16.88} & {\scriptsize{}71.65} & {\scriptsize{}12318}\tabularnewline
{\scriptsize{}OFANet\cite{cai2019once}} & {\scriptsize{}76.10} & {\scriptsize{}17.81} & {\scriptsize{}73.10} & {\scriptsize{}-}\tabularnewline
{\scriptsize{}MNasNet-A1\cite{tan2018mnasnet}} & {\scriptsize{}75.2} & {\scriptsize{}28.65} & {\scriptsize{}125.1} & {\scriptsize{}5642}\tabularnewline
\textbf{\scriptsize{}ET-NAS-E} & \textbf{\scriptsize{}76.87} & {\scriptsize{}25.34} & \textbf{\scriptsize{}61.95} & {\scriptsize{}4922}\tabularnewline
\hline 
{\scriptsize{}EfficientNet-B0\cite{Tan2019}} & {\scriptsize{}77.70} & {\scriptsize{}24.30} & {\scriptsize{}120.29} & {\scriptsize{}7778}\tabularnewline
{\scriptsize{}RegNetY-1.6GF\cite{radosavovic2020designing}} & {\scriptsize{}78.00} & {\scriptsize{}45.59} & {\scriptsize{}170.96} & {\scriptsize{}6338}\tabularnewline
\textbf{\scriptsize{}ET-NAS-F} & \textbf{\scriptsize{}78.80} & \textbf{\scriptsize{}33.83} & \textbf{\scriptsize{}93.04} & \textbf{\scriptsize{}5800}\tabularnewline
\hline 
{\scriptsize{}EfficientNet-B2\cite{Tan2019}} & {\scriptsize{}80.40} & {\scriptsize{}58.78} & {\scriptsize{}277.60} & {\scriptsize{}14258}\tabularnewline
{\scriptsize{}RegNetY-16GF\cite{radosavovic2020designing}} & {\scriptsize{}80.40} & {\scriptsize{}192.78} & {\scriptsize{}677.68} & {\scriptsize{}19258}\tabularnewline
\textbf{\scriptsize{}ET-NAS-G} & {\scriptsize{}80.41} & \textbf{\scriptsize{}53.08} & \textbf{\scriptsize{}133.97} & \textbf{\scriptsize{}8120}\tabularnewline
\textbf{\scriptsize{}ET-NAS-H} & \textbf{\scriptsize{}80.92} & {\scriptsize{}76.80} & {\scriptsize{}193.40} & {\scriptsize{}9140}\tabularnewline
\hline 
{\scriptsize{}EfficientNet-B3\cite{Tan2019}} & {\scriptsize{}81.50} & {\scriptsize{}97.33} & {\scriptsize{}455.86} & {\scriptsize{}22368}\tabularnewline
\textbf{\scriptsize{}ET-NAS-I} & {\scriptsize{}81.38} & \textbf{\scriptsize{}94.60} & \textbf{\scriptsize{}265.13} & \textbf{\scriptsize{}10732}\tabularnewline
\textbf{\scriptsize{}ET-NAS-J} & {\scriptsize{}82.08} & {\scriptsize{}131.92} & {\scriptsize{}370.28} & {\scriptsize{}13774}\tabularnewline
\textbf{\scriptsize{}ET-NAS-L} & \textbf{\scriptsize{}82.65} & {\scriptsize{}191.89} & {\scriptsize{}542.52} & {\scriptsize{}20556}\tabularnewline
\hline 
\end{tabular}{\scriptsize\par}
\par\end{centering}
\begin{centering}
\vspace{1mm}
\par\end{centering}
\begin{centering}
\caption{{\small{}\label{tab:Comparsion-of-ourETNAS}Comparison of our ET-NAS
models and SOTA ImageNet models. Inference time, training step time
and training GPU memory consumption are measured on single Nvidia
V100, with $bs=64$. Our models show a great advantage in terms of
training speed and GPU memory usage.}}
\par\end{centering}
\centering{}\vspace{-1mm}
\end{table}

During the NAS, we directly search on the ImageNet dataset\cite{russakovsky2015imagenet}.
We first search for a group of efficient block structure, then use
those block candidates to conduct the macro-level search. We use a
short training setting to evaluate each architecture. It takes about
1 hour on average for evaluating one architecture for the block-level
search and 6 hours for the macro-level search. Paralleled on GPUs,
it takes about one week on a 64-GPU cluster to conduct the whole search
(5K+1K arch). Implementation details and intermediate results can
be found in the Appendix.

\begin{figure}
\begin{centering}
\vspace{-2mm}
\includegraphics[width=0.8\columnwidth]{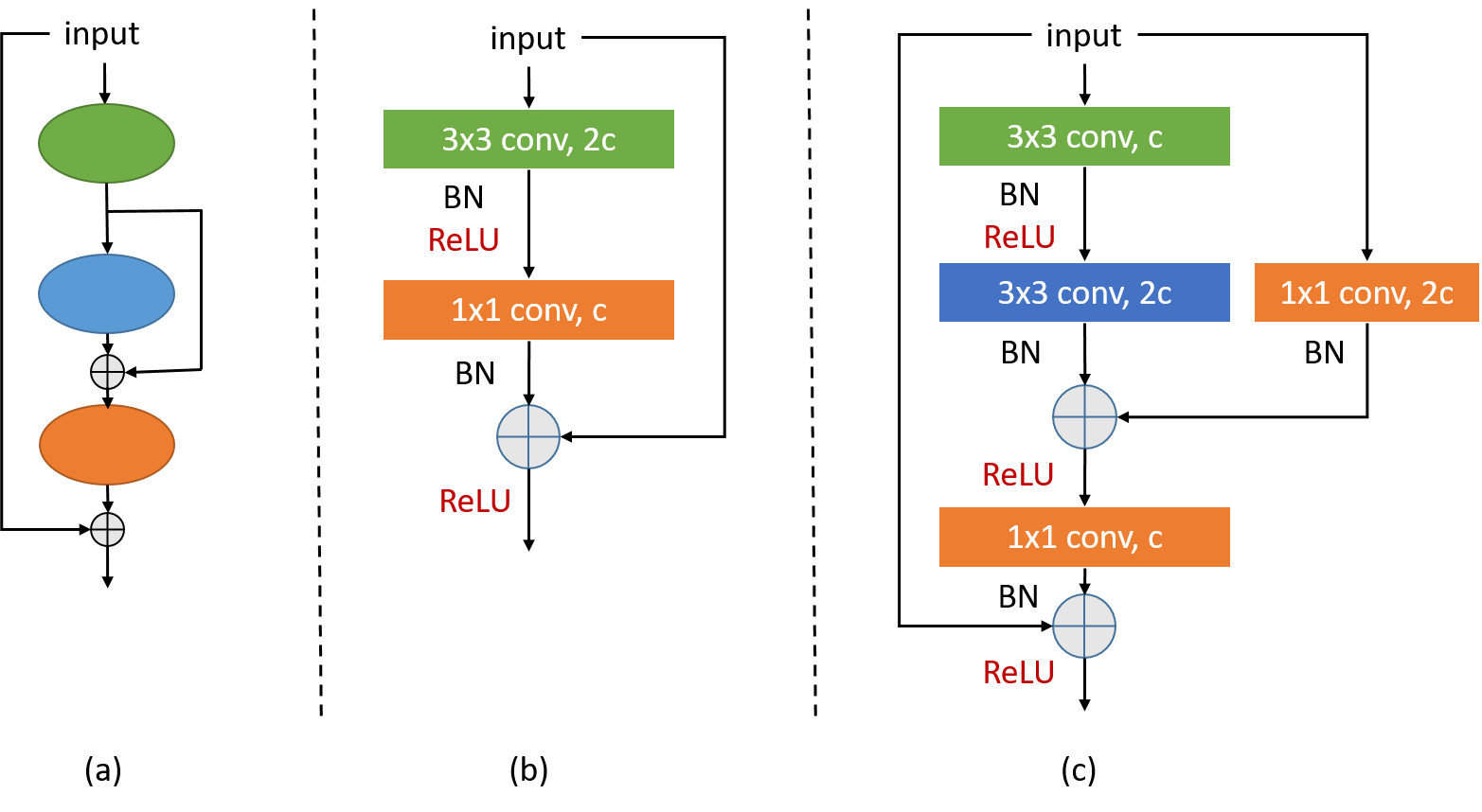}
\par\end{centering}
\begin{centering}
\vspace{-1mm}
\par\end{centering}
\caption{\label{fig:Two-examples-of-block-architecture}Two examples of our
searched block in our ET-NAS. We found that smaller models should
use simpler structures of blocks while bigger models prefer complex
blocks for fast training.}

\centering{}\vspace{-1mm}
\end{figure}

\textbf{Faster Fine-tuning Model Zoo (ET-NAS).} After identifying
the $\mathcal{A}_{i}^{*}$ from our search, we fully train those models
on ImageNet following common practice. Note that all the models including
ET-NAS-L can be easily pretrained on a regular 8-card GPU node since
our model is training-efficient.\textbf{ }We have released our models
for the public to reproduce our results from scratch and let the public
to save their energy/CO2/cost. Due to the length of the paper, we
put the detailed encoding and architectures of the final searched
models in the Appendix. Surprisingly, we found that smaller models
should use simpler structures of blocks while bigger models prefer
complex blocks as shown in Figure \ref{fig:Two-examples-of-block-architecture}.
Comparing our searched backbone to the conventional ResNet/ResNeXt,
we find that early stages in our models are very short which is more
efficient since feature maps in an early stage are very large and
the computational cost is comparably large. This also verified our
findings in preliminary experiments.

\begin{figure*}
\begin{centering}
\vspace{-3mm}
\par\end{centering}
\begin{centering}
\includegraphics[width=0.8\columnwidth]{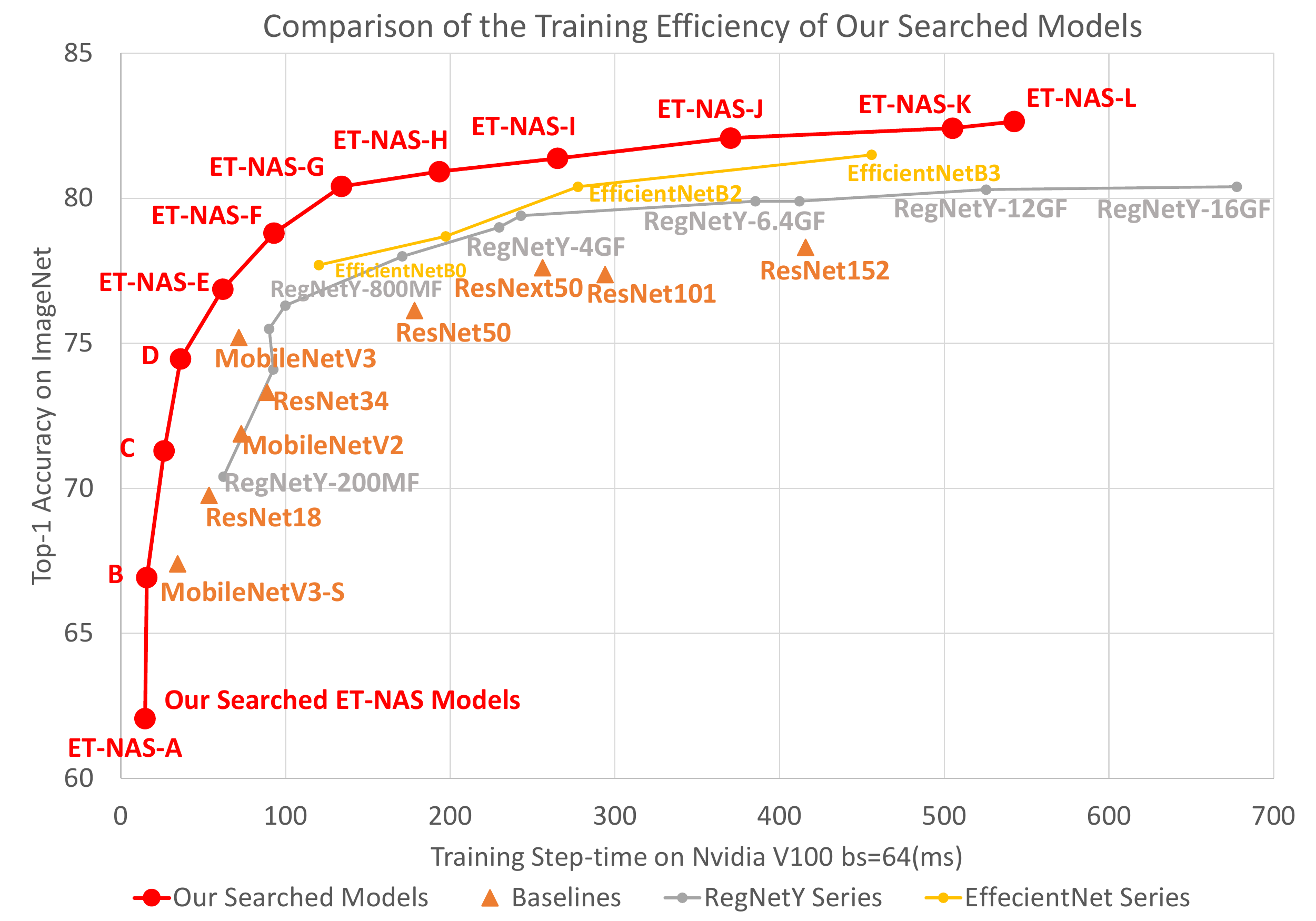}\includegraphics[width=0.8\columnwidth]{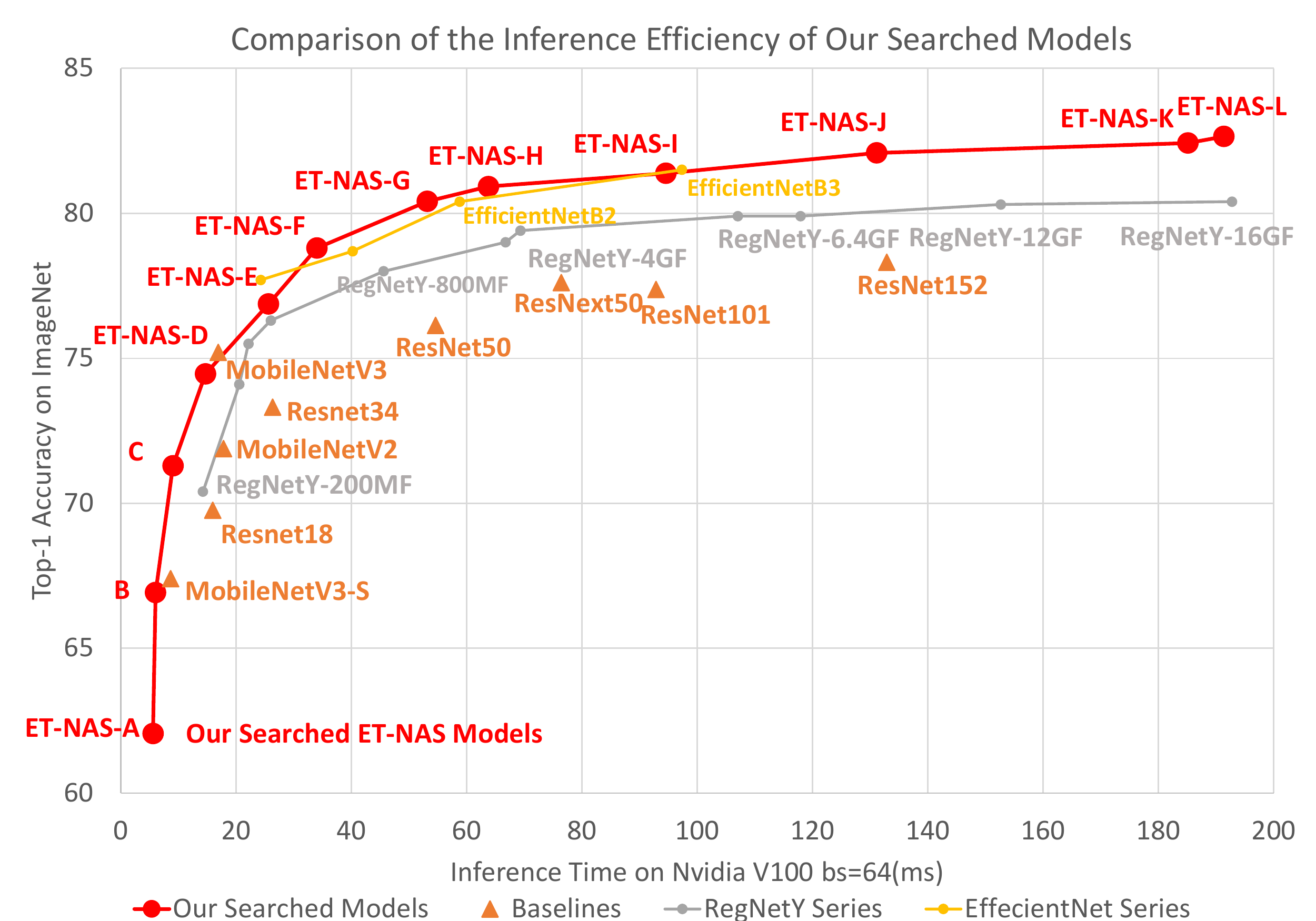}
\par\end{centering}
\begin{centering}
\vspace{-1mm}
\par\end{centering}
\caption{\label{fig:Comparison-of-the-efficiency}Comparison of the training
and inference efficiency of our searched models (ET-NAS) with SOTA
models on ImageNet. Our searched models are faster, e.g., ET-NAS-G
is 6x training faster than RegNetY-16GF, and ET-NAS-I is 1.5x training
faster than EfficientNetB3. Although our models are optimized for
fast training, the inference speed is comparable to EfficientNet and
better than RegNet.}

\centering{}\vspace{-1mm}
\end{figure*}

\textbf{Comparison with the state-of-the-art ImageNet models.} We
compare the training/inference efficiency of our searched ET-NAS with
the SOTA ImageNet models such as MobileNetV3 \cite{howard2019searching},
RegNet series \cite{radosavovic2020designing}, and EfficientNet series
\cite{Tan2019} as shown in Table \ref{tab:Comparsion-of-ourETNAS}
and Figure \ref{fig:Comparison-of-the-efficiency}. Overall, our searched
models outperform other SOTA ImageNet models in terms of training
accuracy and training speed from Figure \ref{fig:Comparison-of-the-efficiency}
(left). Specifically, ET-NAS-G is about 6x training faster than RegNetY-16GF,
and ET-NAS-I is about 1.5x training faster than EfficientNetB3. Our
models are also better than mobile setting models such as MobileNetV2/V3\cite{howard2019searching}
and RegNetY-200MF. Although our model is optimized for fast training,
we also compare the inference speed in Figure \ref{fig:Comparison-of-the-efficiency}(right).
Our models still have a very strong performance in terms of inference
speed, outperforming RegNet series and achieving comparable performance
with EfficientNet. 

\begin{figure}
\begin{centering}
\vspace{-2mm}
\par\end{centering}
\begin{centering}
\includegraphics[width=0.8\columnwidth]{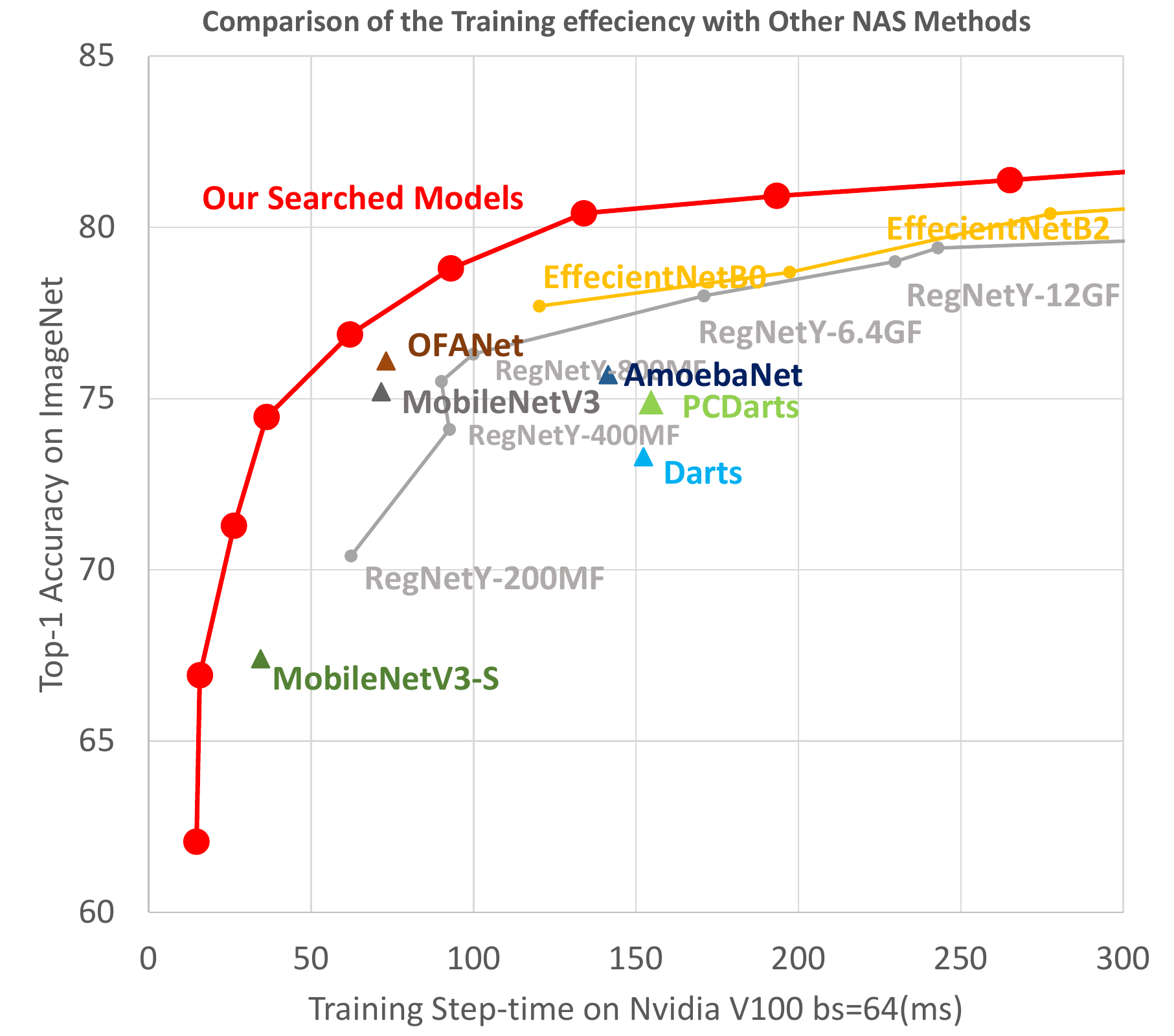}
\par\end{centering}
\centering{}\vspace{-1mm}
\caption{\label{fig:Comparison-NAS}Comparison of the training efficiency of
our searched models (ET-NAS) with 8 other NAS results on ImageNet.
It can be found that our method is more training efficient than some
recent evolution-based NAS methods such as AmoebaNet, OFANet because
of our effective search space.}
\vspace{-1mm}
\end{figure}

\textbf{Comparison with the NAS models.} We also compare our method
with state-of-the-art NAS methods: AmoebaNet\cite{real2019regularized},
OFA\cite{cai2019once}, Darts\cite{liu2018darts}, PCDarts\cite{xu2019pc},
EfficientNet\cite{Tan2019}, RegNet\cite{radosavovic2020designing}
and so on. In Figure \ref{fig:Comparison-NAS}, it can be found that
our searched models are more training efficient than other NAS results,
e.g., some evolution-based NAS methods such as AmoebaNet, OFANet,
and some weight sharing methods such as Darts and AmoebaNet. This
is because of our flexible and effective search space, which considers
both macro and micro-level structure.

\textbf{What makes our network training efficient? }To answer this,
we define an \textit{efficiency score} and conduct a statistical analysis
of different factors for efficient-training (The analysis can be found
in Appendix). We have the following conclusions: a) By observing optimal
$\mathcal{A}_{i}^{*}$, smaller models should use simpler blocks while
bigger models prefer complex blocks. Using the same block structure
for all sizes of models \cite{Tan2019,radosavovic2020designing} may
not be optimal. b) Adding redundant skip-connections which have great
memory access cost will decrease the training efficiency of the model
thus existing topological cell-level search space such as DARTS \cite{liu2018darts},
AmoebaNet \cite{Real2019}, and NASBench101 \cite{dong2019bench}
is not efficient. c) The computation allocation on different stages
is crucially important. Simply increasing depth/width to expand the
model as \cite{Tan2019} may not be optimal will downgrade the performance.
To conclude our novel joint search space contributes most to the training
efficiency.

\subsection{Results for Online Adaptive Predictor $Acc_{P}$ \label{subsec:Online-Training-Regime}}

\textbf{Experimental Settings}. We evaluate our online algorithm based
on ten widely used image classification datasets, that cover various
fine-tuning tasks as shown in Table \ref{tab:Datasets-and-their}.
Five of them (in bold) are chosen to be the online learning training
set (meta-training). 30K samples are collected by continually sampling
a subset of each dataset and fine-tuning with a randomized hyperparameters
on it. Each subset varies from \#. classes and \#. images. The variables
in Section \ref{subsec:Online-Learning-for} are calculated accordingly.
The fine-tuning accuracy is evaluated on the test set. 30K sample
is split into 24K meta-training and 6K meta-validation.

\begin{table}
\begin{centering}
\vspace{-2mm}
\par\end{centering}
\begin{centering}
\renewcommand\arraystretch{0.9}\tabcolsep 0.02in{\scriptsize{}}%
\begin{tabular}{c|cc|cc|cc}
\hline 
 & \multicolumn{2}{c|}{{\scriptsize{}All Cumulative Err.}} & \multicolumn{2}{c|}{{\scriptsize{}Segment 20-40\%}} & \multicolumn{2}{c}{{\scriptsize{}Segment 80-100\%}}\tabularnewline
{\scriptsize{}Models} & {\scriptsize{}MAE} & {\scriptsize{}MSE} & {\scriptsize{}MAE} & {\scriptsize{}MSE} & {\scriptsize{}MAE} & {\scriptsize{}MSE}\tabularnewline
\hline 
{\scriptsize{}Fixed MLP (L=3)} & {\scriptsize{}10.07\%} & {\scriptsize{}1.94\%} & {\scriptsize{}8.99\%} & {\scriptsize{}1.56\%} & {\scriptsize{}7.99\%} & {\scriptsize{}1.21\%}\tabularnewline
{\scriptsize{}Fixed MLP (L=6)} & {\scriptsize{}9.12\%} & {\scriptsize{}1.71\%} & {\scriptsize{}9.03\%} & {\scriptsize{}1.62\%} & {\scriptsize{}7.16\%} & {\scriptsize{}1.04\%}\tabularnewline
{\scriptsize{}Fixed MLP (L=10)} & {\scriptsize{}8.45\%} & {\scriptsize{}1.59\%} & {\scriptsize{}8.46\%} & {\scriptsize{}1.53\%} & {\scriptsize{}6.68\%} & {\scriptsize{}0.96\%}\tabularnewline
{\scriptsize{}Fixed MLP (L=14)} & {\scriptsize{}11.24\%} & {\scriptsize{}2.91\%} & {\scriptsize{}8.34\%} & {\scriptsize{}1.54\%} & {\scriptsize{}4.62\%} & {\scriptsize{}0.46\%}\tabularnewline
\hline 
{\scriptsize{}Ours Adaptive MLP} & \textbf{\scriptsize{}7.51\%} & \textbf{\scriptsize{}1.36\%} & \textbf{\scriptsize{}7.55\%} & \textbf{\scriptsize{}1.11\%} & \textbf{\scriptsize{}3.73\%} & \textbf{\scriptsize{}0.28\%}\tabularnewline
\hline 
\end{tabular}{\scriptsize\par}
\par\end{centering}
\begin{centering}
\vspace{1mm}
\par\end{centering}
\caption{\label{tab:Online-error-rate-1}Online error rate of our method and
fixed MLP. Our adaptive MLP with hedge backpropagation is better in
the online setting of predicting the fine-tuning accuracy.}

\centering{}\vspace{-1mm}
\end{table}

Then an adaptive MLP regression in Eq \ref{eq:adaptive_MLP} are used
to fit the data and predict the $Acc(\mathcal{A}_{i}^{FT},D_{val})$.
We use $L=10$ with 64 units in each hidden layer. We use a learning
rate of $0.01$ and $\beta=0.99$. As baselines, we also compare the
results of using fixed MLP with different layers($L=3,6,10,14$).
MAE (mean absolute error) and MSE (mean square error) are performance
metrics to measure the cumulative error with different segments of
the tasks stream.

\textbf{Comparison of online learning method. }The cumulative error
obtained by all the baselines and the proposed meth od to predict
the fine-tuning accuracy is shown in Table \ref{tab:Online-error-rate}.
Our adaptive MLP with hedge backpropagation is better than fixed MLP
in terms of the cumulative error of the predicted accuracy. Our method
enjoys the benefit from the adaptive depth which allows faster convergence
in the initial stage and strong predictive power in the later stage.

\subsection{\vspace{-1mm}
Final NASOA Results\label{subsec:Final-NASOA-Results}}

\begin{figure*}
\begin{centering}
\vspace{-3mm}
\includegraphics[width=2\columnwidth]{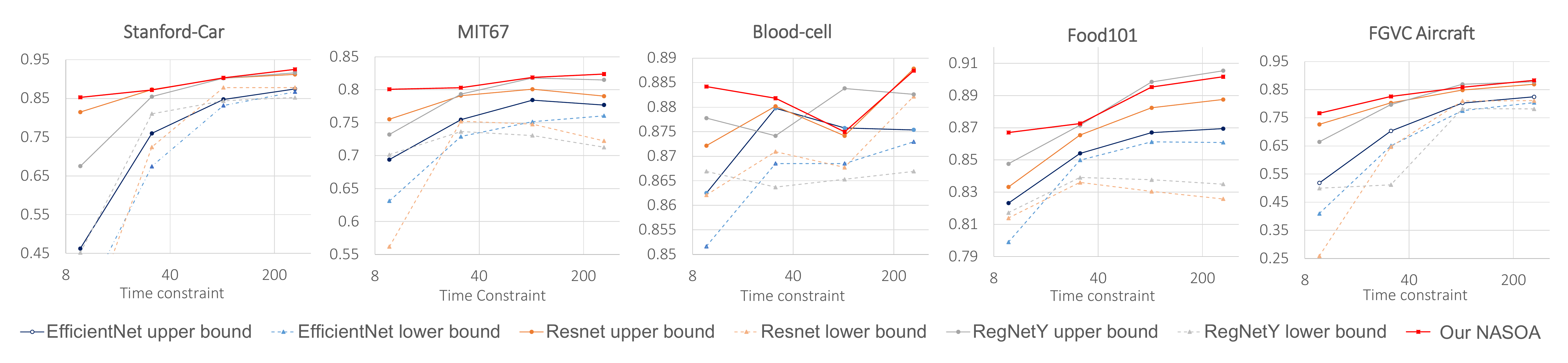}\vspace{-1mm}
\caption{\label{fig:Comparison-the-final-result}Comparison of the final fine-tuning
results under four time constraints for the testing dataset. Red square
lines are the results of our NASOA in one-shot. The dots on the other
solid line are the best performance of all the models in that series
can perform. The model and training regime generated by our NASOA
can outperform the upper bound of other methods in most cases. Our
methods can improve around 2.1\%\textasciitilde 7.4\% accuracy than
the upper bound of RegNet/EfficientNet series on average.}
\par\end{centering}
\centering{}\vspace{-1mm}
\end{figure*}

\begin{table}
\begin{centering}
\renewcommand\arraystretch{0.9}\tabcolsep 0.02in{\tiny{}}%
\begin{tabular}{cccccc}
\hline 
\multicolumn{2}{r}{{\scriptsize{}Methods \hspace{8mm}Search Cost}} & {\scriptsize{}Aircrafts} & {\scriptsize{}MIT67} & {\scriptsize{}Sf-Car} & {\scriptsize{}Sf-Dog}\tabularnewline
\hline 
{\scriptsize{}Random Search (HPO only)} & {\scriptsize{}x40} & {\scriptsize{}63.07\%} & {\scriptsize{}75.60\%} & {\scriptsize{}67.47\%} & {\scriptsize{}86.25\%}\tabularnewline
{\scriptsize{}BOHB (HPO only)} & {\scriptsize{}x40} & {\scriptsize{}72.70\%} & {\scriptsize{}77.61\%} & {\scriptsize{}70.94\%} & {\scriptsize{}87.41\%}\tabularnewline
\hline 
{\scriptsize{}Random Search} & {\scriptsize{}x40} & {\scriptsize{}81.07\%} & {\scriptsize{}79.93\%} & {\scriptsize{}88.99\%} & {\scriptsize{}89.06\%}\tabularnewline
{\scriptsize{}BOHB} & {\scriptsize{}x40} & \textbf{\scriptsize{}82.34\%} & \textbf{\scriptsize{}79.85\%} & \textbf{\scriptsize{}89.01\%} & \textbf{\scriptsize{}89.49\%}\tabularnewline
\hline 
{\scriptsize{}Our $Z_{oo}$ with Random Search} & {\scriptsize{}x40} & {\scriptsize{}83.71\%} & {\scriptsize{}80.97\%} & {\scriptsize{}87.84\%} & {\scriptsize{}92.75\%}\tabularnewline
{\scriptsize{}Our $Z_{oo}$ with BOHB} & {\scriptsize{}x40} & {\scriptsize{}84.67\%} & {\scriptsize{}82.34\%} & {\scriptsize{}89.03\%} & {\scriptsize{}93.74\%}\tabularnewline
\hline 
{\scriptsize{}Our OA only} & {\scriptsize{}x1} & {\scriptsize{}81.22\%} & {\scriptsize{}79.33\%} & {\scriptsize{}84.56\%} & {\scriptsize{}89.70\%}\tabularnewline
{\scriptsize{}Our $Z_{oo}$ with Fixed MLP (Offline)} & {\scriptsize{}x1} & {\scriptsize{}81.31\%} & {\scriptsize{}75.97\%} & {\scriptsize{}88.81\%} & {\scriptsize{}88.58\%}\tabularnewline
\textbf{\scriptsize{}Our final NASOA} & \textbf{\scriptsize{}x1} & \textbf{\scriptsize{}82.54\%} & \textbf{\scriptsize{}80.30\%} & \textbf{\scriptsize{}88.20\%} & \textbf{\scriptsize{}92.30\%}\tabularnewline
\hline 
\end{tabular}{\tiny\par}
\par\end{centering}
\caption{\label{tab:Online-error-rate}Comparison of the final NASOA results
with other HPO methods. ``HPO only'' means only optimizing the hyperparameters
with RegNetY-16GF. Other HPO methods optimize both selecting hyperparameters
and model from RegNet series models. ``OA only'' is our online schedule
generator with RegNet series models. ``Our $Z_{oo}$'' means using
our models zoo to find suitable model. \textquotedblleft Fixed MLP
Predictor\textquotedblright{} is the offline baseline with fixed MLP
predictor. ``Our NASOA'' is our whole pipeline with both the model
zoo and online adaptive scheduler. Without additional search cost
(x40), NASOA can reach similar performance of BOHB.}

\centering{}\vspace{-1mm}
\end{table}

To evaluate the performance of our whole NASOA, we select four time
constraints on the testing datasets and use $Acc_{P}(.)$ to test
the fine-tuning accuracy. The testing datasets are MIT67, Food101,
Aircrafts, Blood-cell, and Stanford-Car. The shortest/longest time
constraint are the time of fine-tuning 10/50 epochs for ResNet18/ResNet101.
The rest are equally divided into the log-space. For our NASOA, we
generate the fine-tuning schedules by maximizing the predicted accuracy
in Eq \ref{eq:regimegenearator}. We also conduct fine-tuning on various
candidates of baselines such as ResNet (R18 to R152), RegNet (200MF
to 16GF), and EfficientNet (B0 to B3) with the default hyperparameter
setting in \cite{li2020rethinking}.

\textbf{Comparison of the final fine-tuning results with the SOTA
networks. }We plot the time versus accuracy comparison in Figure \ref{fig:Comparison-the-final-result}.
As can be seen, the model and training regime generated by our NASOA
can outperform the upper bound of other methods in most cases. On
average, our methods can improve around 2.1\%/7.4\% accuracy than
the best model of RegNet/EfficientNet series under various time constraints
and tasks. It is noteworthy that our NASOA performs better especially
in the case of short time constraint, which demonstrates that our
schedule generator is capable provide both efficient and effective
regimes. 

\begin{table}
\begin{centering}
\vspace{-1mm}
\renewcommand\arraystretch{1}\tabcolsep 0.05in{\scriptsize{}}%
\begin{tabular}[b]{l|c|c|c|c|l|l}
\hline 
\multirow{2}{*}{\textbf{\scriptsize{}Methods}} & {\scriptsize{}Fixed} & {\scriptsize{}Existing} & {\scriptsize{}NAS} & {\scriptsize{}Adaptive} & \textbf{\scriptsize{}Comp} & \textbf{\scriptsize{}Avg. Finetune}\tabularnewline
 & {\scriptsize{}Model} & {\scriptsize{}Models} & {\scriptsize{}Models} & {\scriptsize{}Scheduler} & \textbf{\scriptsize{}Cost} & \textbf{\scriptsize{}Accuracy}\tabularnewline
\hline 
{\scriptsize{}BOHB\cite{falkner2018bohb}} & {\scriptsize{}$\checked$} &  &  &  & {\scriptsize{}x40} & {\scriptsize{}77.2\%}\tabularnewline
\multirow{2}{*}{{\scriptsize{}+ Our $Z_{oo}$}} &  &  & \multirow{2}{*}{{\scriptsize{}$\checked$}} &  & \multirow{2}{*}{{\scriptsize{}x40}\textcolor{blue}{\scriptsize{}$^{-0\times}$}} & \multirow{2}{*}{{\scriptsize{}87.5\%}\textcolor{blue}{\scriptsize{}$^{+10.3\%}$}}\tabularnewline
 &  &  &  &  &  & \tabularnewline
\multirow{2}{*}{{\scriptsize{}Our OA}} &  & \multirow{2}{*}{{\scriptsize{}$\checked$}} &  & \multirow{2}{*}{{\scriptsize{}$\checked$}} & \multirow{2}{*}{{\scriptsize{}x1}\textcolor{blue}{\scriptsize{}$^{-40\times}$}} & \multirow{2}{*}{{\scriptsize{}83.7\%}\textcolor{blue}{\scriptsize{}$^{+6.5\%}$}}\tabularnewline
 &  &  &  &  &  & \tabularnewline
\multirow{2}{*}{{\scriptsize{}NASOA}} &  &  & \multirow{2}{*}{{\scriptsize{}$\checked$}} & \multirow{2}{*}{{\scriptsize{}$\checked$}} & \multirow{2}{*}{{\scriptsize{}x1}\textcolor{blue}{\scriptsize{}$^{-40\times}$}} & \multirow{2}{*}{{\scriptsize{}85.8\%}\textcolor{blue}{\scriptsize{}$^{+8.6\%}$}}\tabularnewline
 &  &  &  &  &  & \tabularnewline
\hline 
\end{tabular}{\scriptsize\par}
\par\end{centering}
\centering{}\caption{\label{tab:Ablative-study} Ablative study. We calculates the average
fine-tuning accuracy over the test datasets.}
\vspace{-2mm}
\end{table}

\textbf{Comparison of the final fine-tuning results with the HPO methods.}
In Table \ref{tab:Online-error-rate}, we compare our method with
the HPO methods which optimizing the hyperparameters and picking up
models in ResNet, RegNetY, and EfficientNet series. ``HPO only''
means the method only optimizes the hyperparameters with a fixed model
RegNetY-16GF.``OA only'' is our online schedule generator with RegNet
series models. ``Our $Z_{oo}$'' means using our ET models zoo to
find suitable model. \textquotedblleft Fixed MLP Predictor\textquotedblright{}
is the offline baseline with fixed MLP predictor (L=10) with our model
zoo. ``Our NASOA'' is the our whole pipeline with both training
efficient model zoo and online adaptive scheduler. Comparing to the
offline baseline with our NASOA, our online adaption module can boost
the average performance by 2.17\%. It can also be found that our method
can save up to 40x computational cost compared to HPO methods while
reaching similar performance. With more computational budget, our
model zoo with BOHB search can reach even higher accuracy (+avg. 10.28\%).

\textbf{Ablative interpretation of performance superiority. }Table
\ref{tab:Ablative-study} calculates the average fine-tuning accuracy
over tasks. Our NAS model zoo can greatly increase the fine-tuning
average accuracy from $77.17$\% to $87.45$\%, which is the main
contribution of the performance superiority. Using our online adaptive
scheduler instead of BOHB can significantly reduce the computational
cost (-40x).

\section{\vspace{-2mm}
Conclusion}

We propose the first efficient task-oriented fine-tuning framework
aiming at saving the resources for GPU clusters and cloud computing.
The joint NAS and online adaption strategy achieves much better fine-tuning
results in terms of both accuracy and speed. The searched architectures
are more training-efficient than very strong baselines. We also theoretically
prove our online model is better than fixed-depth model. Our experiments
on multiple datasets show our NASOA achieves 40x speed-up comparing
to BOHB. Generalization to more tasks such as detection and segmentation
can be considered for future work.

{\small{}\bibliographystyle{ieee_fullname}
\bibliography{Know_network}
}{\small\par}

\newpage{}

\part*{Appendix}

\section{Preliminary experiments\label{sec:Preliminary-experiments}}

\subsection{Experiments Settings}

The preliminary experiments aim at figuring out what kinds of factors
impact the speed and accuracy of fine-tuning. We fine-tune several
ImageNet pretrained backbones on various datasets as shown in Table
\ref{tab:Preliminary-Experiment-1:} (right) and exam different settings
of hyperparameters by a grid search such as: learning rate (0.0001,
0.001, 0.01, 0.1), frozen stages (-1,0,1,2,3), and frozen BN (-1,0,1,2,3).
Frozen stages/frozen BN$=k$ means $1$ to $k$th stage's parameters/BN
statistics are not updated during training. The training settings
most follow \cite{li2020rethinking} and we report the Top-1 validation
accuracy and training time. Its detailed experiment settings are hyperparameters
are listed as follows:

\begin{table*}
\begin{centering}
\vspace{-2mm}
\par\end{centering}
\begin{centering}
\renewcommand\arraystretch{1}\tabcolsep 0.04in{\footnotesize{}}%
\begin{tabular}{cccccccc}
\hline 
\multirow{2}{*}{{\footnotesize{}Dataset}} & \multirow{2}{*}{{\footnotesize{}Method}} & \multicolumn{2}{c}{{\footnotesize{}ResNet18}} & \multicolumn{2}{c}{{\footnotesize{}ResNet34}} & \multicolumn{2}{c}{{\footnotesize{}ResNet50}}\tabularnewline
\cline{3-8} \cline{4-8} \cline{5-8} \cline{6-8} \cline{7-8} \cline{8-8} 
 &  & {\footnotesize{}Acc.} & {\footnotesize{}Time (min)} & {\footnotesize{}Acc.} & {\footnotesize{}Time (min)} & {\footnotesize{}Acc.} & {\footnotesize{}Time (min)}\tabularnewline
\hline 
\multirow{2}{*}{{\footnotesize{}Flowers102}} & {\footnotesize{}From Scratch} & {\footnotesize{}94.4\%} & {\footnotesize{}11} & {\footnotesize{}93.8\%} & {\footnotesize{}19} & {\footnotesize{}90.7\%} & {\footnotesize{}38}\tabularnewline
 & {\footnotesize{}Fine-tuning} & \textbf{\footnotesize{}98.3\%} & \textbf{\footnotesize{}7} & \textbf{\footnotesize{}98.5\%} & \textbf{\footnotesize{}11} & \textbf{\footnotesize{}98.8\%} & \textbf{\footnotesize{}22}\tabularnewline
\hline 
\multirow{2}{*}{{\footnotesize{}Stanford-Car}} & {\footnotesize{}From Scratch} & {\footnotesize{}80.6\%} & {\footnotesize{}14} & {\footnotesize{}81.9\%} & {\footnotesize{}24} & {\footnotesize{}81.5\%} & {\footnotesize{}47}\tabularnewline
 & {\footnotesize{}Fine-tuning} & \textbf{\footnotesize{}87.6\%} & \textbf{\footnotesize{}8} & \textbf{\footnotesize{}89.6\%} & \textbf{\footnotesize{}14} & \textbf{\footnotesize{}91.1\%} & \textbf{\footnotesize{}28}\tabularnewline
\hline 
\multirow{2}{*}{{\footnotesize{}CUB-Birds}} & {\footnotesize{}From Scratch} & {\footnotesize{}51.6\%} & {\footnotesize{}11} & {\footnotesize{}53.6\%} & {\footnotesize{}18} & {\footnotesize{}44.6\%} & {\footnotesize{}35}\tabularnewline
 & {\footnotesize{}Fine-tuning} & \textbf{\footnotesize{}69.2\%} & \textbf{\footnotesize{}6} & \textbf{\footnotesize{}71.9\%} & \textbf{\footnotesize{}10} & \textbf{\footnotesize{}74.9\%} & \textbf{\footnotesize{}21}\tabularnewline
\hline 
\multirow{2}{*}{{\footnotesize{}MIT67}} & {\footnotesize{}From Scratch} & {\footnotesize{}67.8\%} & {\footnotesize{}22} & {\footnotesize{}69.2\%} & {\footnotesize{}37} & {\footnotesize{}66.0\%} & {\footnotesize{}73}\tabularnewline
 & {\footnotesize{}Fine-tuning} & \textbf{\footnotesize{}76.4\%} & \textbf{\footnotesize{}13} & \textbf{\footnotesize{}76.9\%} & \textbf{\footnotesize{}21} & \textbf{\footnotesize{}78.1\%} & \textbf{\footnotesize{}43}\tabularnewline
\hline 
\multirow{2}{*}{{\footnotesize{}Stanford-Dog}} & {\footnotesize{}From Scratch} & {\footnotesize{}60.6\%} & {\footnotesize{}21} & {\footnotesize{}62.6\%} & {\footnotesize{}35} & {\footnotesize{}55.2\%} & {\footnotesize{}70}\tabularnewline
 & {\footnotesize{}Fine-tuning} & \textbf{\footnotesize{}69.7\%} & \textbf{\footnotesize{}12} & \textbf{\footnotesize{}73.3\%} & \textbf{\footnotesize{}20} & \textbf{\footnotesize{}75.0\%} & \textbf{\footnotesize{}41}\tabularnewline
\hline 
\multirow{2}{*}{{\footnotesize{}Caltech101}} & {\footnotesize{}From Scratch} & {\footnotesize{}82.5\%} & {\footnotesize{}12} & {\footnotesize{}78.1\%} & {\footnotesize{}20} & {\footnotesize{}75.7\%} & {\footnotesize{}40}\tabularnewline
 & {\footnotesize{}Fine-tuning} & \textbf{\footnotesize{}90.8\%} & \textbf{\footnotesize{}7} & \textbf{\footnotesize{}91.8\%} & \textbf{\footnotesize{}12} & \textbf{\footnotesize{}91.8\%} & \textbf{\footnotesize{}23}\tabularnewline
\hline 
\end{tabular}{\footnotesize\par}
\par\end{centering}
\begin{centering}
\vspace{1mm}
\par\end{centering}
\begin{centering}
\caption{\label{tab:Preliminary-Experiment-1:}{\small{}Comparison of Top-1
accuracy and training time (min) on different datasets. Comparing
to training from scratch, fine-tuning shows superior results in terms
of both accuracy and training time.}}
\par\end{centering}
\vspace{-1mm}
\end{table*}

\textbf{Comparing fine-tuning and training from scratch}. We use ResNet
series (R-18 to R-50) to evaluate the effect of fine-tuning and training
from scratch. Following \cite{li2020rethinking}, we train networks
on Flowers102, CUB-Birds, MIT67 and Caltech101 datasets for 600 epochs
for training from scratch and 350 epochs for fine-tuning to ensure
all models converge on all datasets. We use SGD optimizer with an
initial learning rate 0.01, weight decay 1e-4, momentum 0.9. The learning
rate is decreased by factor 10 at 400 and 550 epoch for training from
scratch and 150, 250 epoch for fine-tuning.

\textbf{Optimal learning rate and frozen stage}. We perform a simple
grid search on Flowers102, Stanford-Car, CUB-Birds, MIT67, Stanford-Dog,
and Caltech101 datasets with ResNet50 to find optimal learning rate
and frozen stage on different datasets with the default fine-tune
setting in \cite{li2020rethinking}. The hyperparameters ranges are:
learning rate (0.1, 0.01, 0.001, 0.0001), frozen stage (-1, 0, 1,
2, 3).

\begin{table*}
\begin{centering}
\vspace{-1mm}
\par\end{centering}
\begin{centering}
\vspace{1mm}
{\tiny{}\tabcolsep 0.04in}{\footnotesize{}}%
\begin{tabular}{cccccc}
\hline 
\multirow{2}{*}{{\footnotesize{}DataSet}} & \multirow{2}{*}{{\footnotesize{}\#.Class}} & \multirow{2}{*}{{\footnotesize{}\#.Images}} & {\footnotesize{}Optimal} & {\footnotesize{}Optimal} & \multirow{2}{*}{{\footnotesize{}Best Acc.}}\tabularnewline
 &  &  & {\footnotesize{}LR} & {\footnotesize{}Frozen Stage} & \tabularnewline
\hline 
{\footnotesize{}Flowers102} & {\footnotesize{}102} & {\footnotesize{}8K} & {\footnotesize{}0.01} & {\footnotesize{}0} & {\footnotesize{}99.3\%}\tabularnewline
{\footnotesize{}Stanford-Car} & {\footnotesize{}196} & {\footnotesize{}16K} & {\footnotesize{}0.1} & {\footnotesize{}1} & {\footnotesize{}91.8\%}\tabularnewline
{\footnotesize{}CUB-Birds} & {\footnotesize{}200} & {\footnotesize{}12K} & {\footnotesize{}0.01} & {\footnotesize{}2} & {\footnotesize{}81.3\%}\tabularnewline
{\footnotesize{}MIT67} & {\footnotesize{}67} & {\footnotesize{}16K} & {\footnotesize{}0.01} & {\footnotesize{}-1} & {\footnotesize{}80.8\%}\tabularnewline
{\footnotesize{}Stanford-Dog} & {\footnotesize{}120} & {\footnotesize{}21K} & {\footnotesize{}0.01} & {\footnotesize{}3} & {\footnotesize{}83.7\%}\tabularnewline
{\footnotesize{}Caltech101} & {\footnotesize{}101} & {\footnotesize{}8K} & {\footnotesize{}0.001} & {\footnotesize{}-1} & {\footnotesize{}96.4\%}\tabularnewline
{\footnotesize{}Caltech256} & {\footnotesize{}257} & {\footnotesize{}31K} & {\footnotesize{}0.01} & {\footnotesize{}3} & {\footnotesize{}85.6\%}\tabularnewline
\hline 
\end{tabular}{\footnotesize\par}
\par\end{centering}
\begin{centering}
\vspace{1mm}
\par\end{centering}
\begin{centering}
\caption{\label{tab:Preliminary-Experiment-2}{\small{}Fine-tuning on R50,
the optimal learning rate and optimal frozen stage found by grid search
are different and should be optimized individually.}}
\par\end{centering}
\centering{}\vspace{-1mm}
\end{table*}

\begin{figure*}
\begin{centering}
\vspace{-1mm}
\includegraphics[width=2\columnwidth]{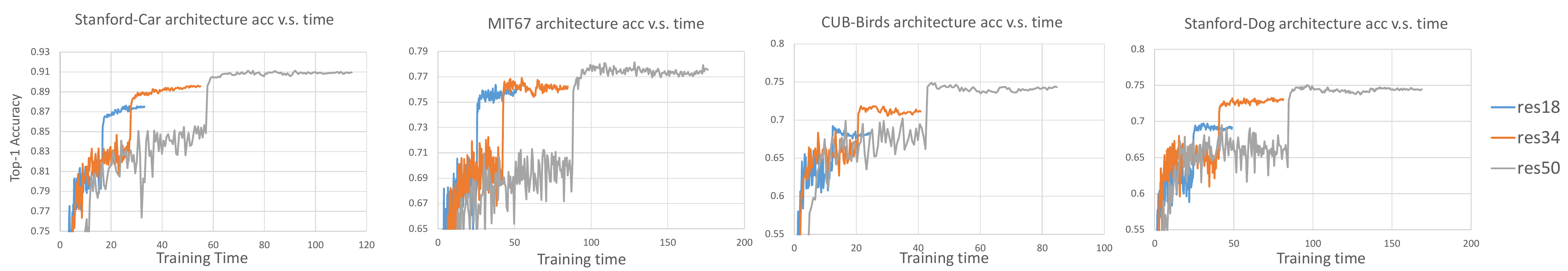}\vspace{1mm}
\par\end{centering}
\caption{\label{fig:Fine-tune-results-along}Fine-tune results along time with
various networks on these datasets. It can be seen that if the time
constraints is short, we should choose a smaller network.}
\vspace{-1mm}
\end{figure*}

\textbf{Comparing different frozen stages and networks along time}.
We fix different stages of ResNet50 to analyze the influence of different
frozen stages to the accuracy and along the training time on Flowers102,
Stanford-Car, CUB-Birds, MIT67, Stanford-Dog, and Caltech101 datasets.
We pick the training curves on CUB-Birds and Caltech101 to in the
main text of this paper. We also compare the fine-tune results along
time with various networks on these datasets as shown in Figure \ref{fig:Fine-tune-results-along}.
On Caltech101, ResNet50 dominates the training curve at the very beginning.
However, on other datasets, ResNet18 and ResNet34 can perform better
then ResNet50 when the training time is short.

\begin{figure*}
\begin{centering}
\vspace{-1mm}
\par\end{centering}
\begin{centering}
\includegraphics[width=2\columnwidth]{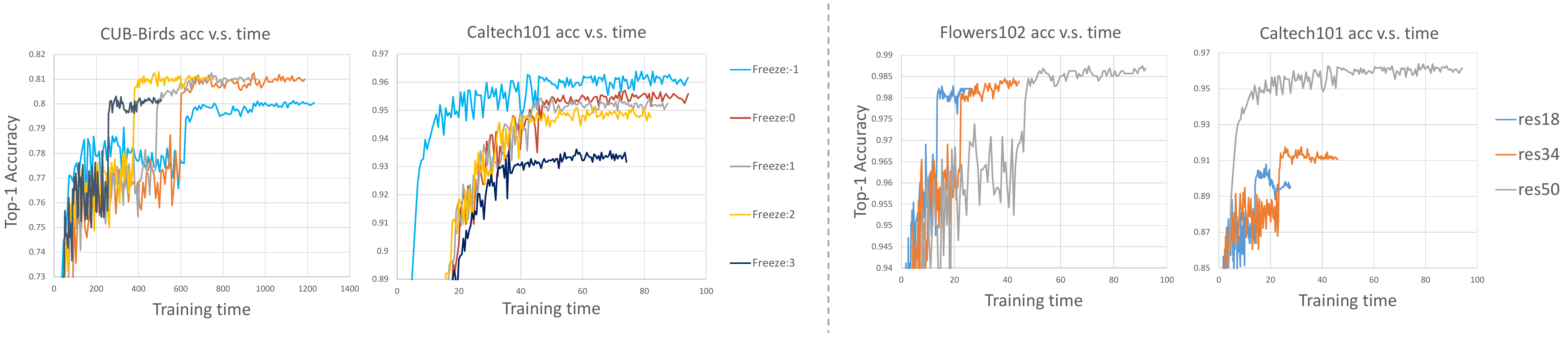}
\par\end{centering}
\begin{centering}
\vspace{1mm}
\par\end{centering}
\caption{\label{fig:Preliminary_Exp_2}{\small{}(Left) Fine-tuning ResNet101
with different weight-frozen stages. ``Freeze: k'' means 0 to k
stage's parameters are not updated during training. The number of
frozen stage will effect both training time and accuracy. Its optimal
frozen setting varies with datasets. (Right) Comparison of accuracy/time
different fine-tuning models. Different models should be selected
upon the request of different datasets and training constraints.}}
\vspace{-1mm}
\end{figure*}

\subsection{Findings of the preliminary experiments}

With those preliminary experiments, we summarize our findings as follows.
Some of the findings are also verified by some existing works.

\begin{itemize}[leftmargin=*]

\item \textbf{Fine-tuning performs always better than training from
scratch.} As shown in Table \ref{tab:Preliminary-Experiment-1:},
fine-tuning shows superior results than training from scratch in terms
of both accuracy and training time for all the datasets. This finding
is also verified by \cite{kornblith2019better}. Thus, fine-tuning
is the most common way to train a new dataset and our framework can
be generalized to applications.

\item \textbf{We should the optimize learning rate and frozen stage
for each dataset.} From Table \ref{tab:Preliminary-Experiment-2},
it seems that the optimal learning rate and optimal frozen stage found
by grid search are different for various datasets. Figure \ref{fig:Preliminary_Exp_2}also
shows that the number of the frozen stages will affect both training
time and final accuracy. \cite{guo2019spottune} also showed that
frozen different stages are crucial for fine-tuning task. Those two
hyperparameters should be optimized for different datasets.

\item \textbf{Model matters most. Suitable model should be selected
according to the task and time constraints.} Figure \ref{fig:Preliminary_Exp_2}
(right) suggests that always choosing the biggest model to fine-tune
may not be an optimal choice, smaller model can be better than the
bigger model if the training time is limited. On the other hand, it
is also important to consider the training efficiency of the model
since a better model can be converged faster by a limited GPU budget.
For example, Figure \ref{fig:Fine-tune-results-along} shows that
if the time constraint is short, we should choose a smaller network
i.e. ResNet18 here. Thus, it is urgent to construct a training-efficient
model zoo.

\item \textbf{BN running statistics should not be frozen during fine-tuning.}
We found that frozen BN has a very limited effect on the training
time (less than $\pm5\%$), while not freezing BN will lead to better
results in almost all the datasets. Thus, BN is not frozen all experiments
for our NASOA.

\end{itemize}

\section{Details of the Offline NAS \label{sec:Details-of-theNAS}}

\subsection{Search Space Encodings\label{subsec:Search-Space-Encodings}}

The search space of our architectures is composed of block and macro
levels, where the former decides what a block is composed of, such
as operators, number of channels, and skip connections, while the
latter is concerned about how to combine the block into a whole network,
e.g., when to do down-sampling, and where to change the number of
channels.

\subsubsection{Block-level architecture}

\begin{figure}
\begin{centering}
\includegraphics[width=1\columnwidth]{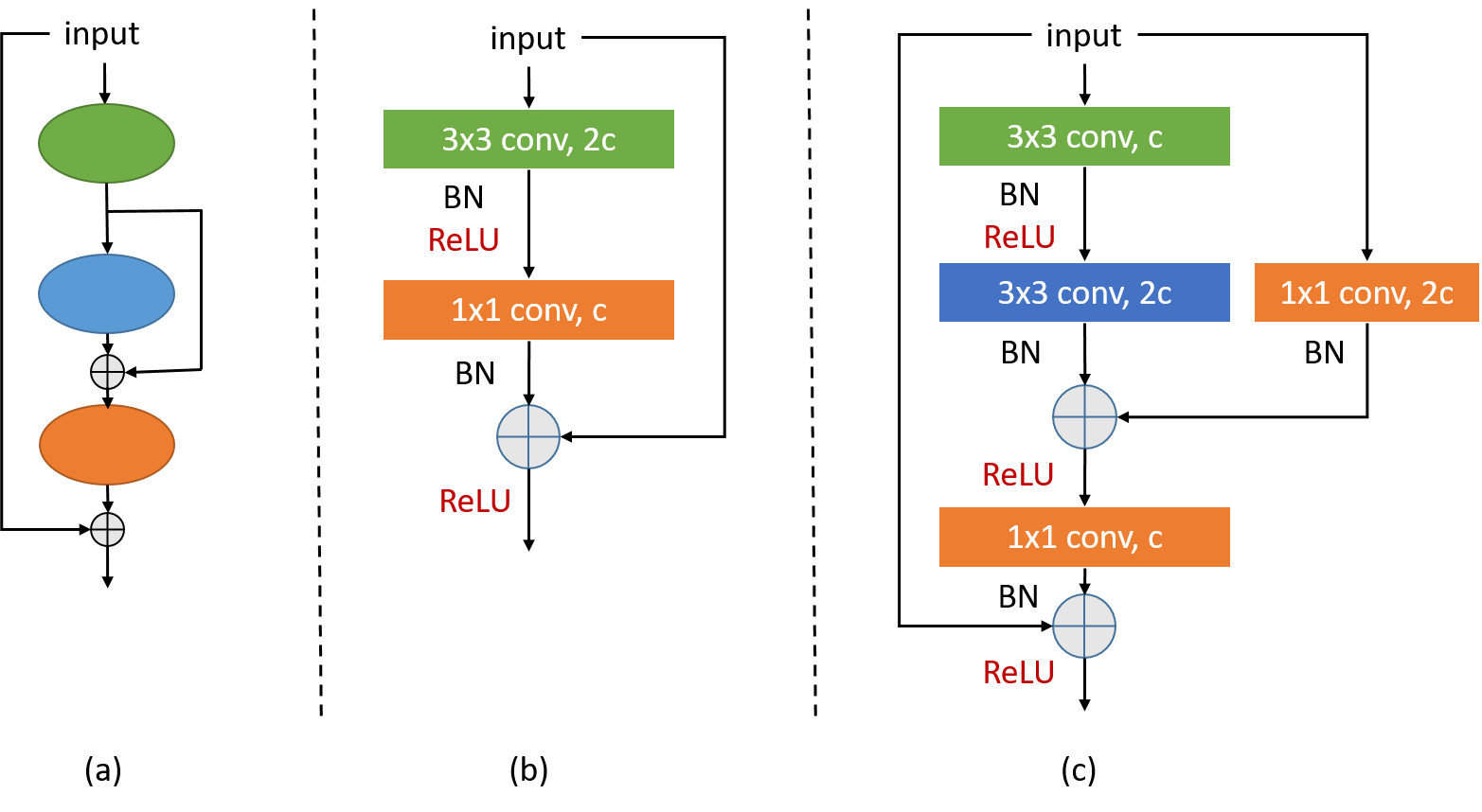}
\par\end{centering}
\caption{\label{figure:block_examples}Block structure and two block samples.
(a) shows a three-node graph. (b) is an example with encoding ``031-'',
and (c) is ``02031-a02''.}

\end{figure}

\textbf{Block-level design.} A block consists of at most three operators,
each of which is divided into 5 species and has 5 different number
of output channels. Each kind of operator is denoted by an op number,
and the output channel of the operator is decided by the ratio between
it and that of the current block. Details are shown in Table \ref{table:micro_arch}.
By default, there is a skip connection between the input and output
of the block, which sums their values up. In addition to that, at
most 3 other skip connections are contained in a block, which either
adds or concatenates the values between them. Each operation is followed
by a batch normalization layer, and after all the skip connections
are calculated, a ReLU layer is triggered.

\textbf{Block-level encoding.} The encoding of each block-level architecture
is composed of two parts separated by `-', which considers the operators
and skip connections respectively.

For the first part (operators part), each operator is represented
by two numbers: op number and ratio number (shown in Table \ref{table:micro_arch}).
As the output channel of the last operation always equals to that
of the current block, the ratio number of this operator is removed.
Therefore, the encoding of a block with $n$ operators always has
length $2n-1$ for the first part of block-level encoding.

For the second part (skip connections part), every skip connection
consists of one letter for addition (`a') / concatenation (`c'), and
two numbers for place indices. $n$ operators separate the block to
$n+1$ parts, which are indexed with $0,1,2,\dots,n$. Thus `a01'
means summing up the value before and after the first operator. Since
the skip connection between the beginning and end of the block always
exists, it is not shown in the encoding. Thus this part has length
$3k-3$ (possibly $0$) when there is $k$ skip connections. Some
of the encoding examples are shown in Figure \ref{figure:block_examples}.

\begin{table}[b]
\begin{centering}
\vspace{1mm}
\begin{tabular}{clcl}
\toprule 
op number & operator(Input $c$) & ratio number & ratios\tabularnewline
\midrule 
0 & conv3x3 & 0 & $\times1/4$\tabularnewline
1 & conv1x1 & 1 & $\times1/2$\tabularnewline
2 & conv3x3, group=2 & 2 & $\times1$\tabularnewline
3 & conv3x3, group=4 & 3 & $\times2$\tabularnewline
4 & conv3x3, group=$c$ & 4 & $\times4$\tabularnewline
\bottomrule
\end{tabular}\label{table:micro_arch}
\par\end{centering}
\begin{centering}
\vspace{1mm}
\par\end{centering}
\centering{}\caption{The operations and channel changing ratios considered in our paper.
Encoding for operators and ratios. $c$ stands for the channels of
the current block.}
\end{table}

\subsubsection{Macro-level architecture}

\textbf{Macro-level design. }We only consider networks with exactly
4 stages. The first block of each stage (except Stage 1) reduces the
resolution of both width and height by half, where the stride 2 is
added to the first operator that is not conv1x1. Other blocks do not
change the resolution. One block's output channel is either the same,
or an integer multiple of the input channel.

\textbf{Macro-level encoding. }The 4 stages are divided apart by 3
`-' signs. For every stage, each block is represented by an integer,
which shows the ratio between output and input channel for this block.

\subsubsection{Encoding as a whole}

\begin{table}
\begin{centering}
\vspace{1mm}
\begin{tabular}{r|l}
\hline 
\textbf{model} & \textbf{encoding}\tabularnewline
\hline 
ResNet18 & 020-\_64\_11-21-21-21\tabularnewline
\hline 
ResNet34 & 020-\_64\_111-2111-211111-211\tabularnewline
\hline 
ResNet50 & 10001-\_64\_411-2111-211111-211\tabularnewline
\hline 
Wide ResNet50 & 11011-\_64\_411-2111-211111-211\tabularnewline
\hline 
\end{tabular}
\par\end{centering}
\begin{centering}
\vspace{1mm}
\par\end{centering}
\begin{centering}
\label{table:encoding_example}
\par\end{centering}
\begin{centering}
\caption{ResNets and Wide ResNets are represented by our encoding scheme. Basic
Block is represented as `020-', as the two operators are both conv3x3
(denoted as `0'), and the output channel of the first operator equals
to that of the block output (represented as `2'), and no other skip
connection except the one connecting input and output; the macro-arch
of ResNet 18 is encoded as `11-21-21-21', as each stage contains two
blocks, where the first block in Stage 2, 3, 4 doubles the number
of channels.}
\par\end{centering}
\centering{}\vspace{-1mm}
\end{table}

Thus the whole backbone can be encoded by simply concatenating the
block and macro encoding. The encoding of the whole network is formatted
as: {\scriptsize{}
\[
\{Block\_ENCODING\}\_\{First\_CHANNEL\}\_\{Macro\_ENCODING\}
\]
}Some common architectures, including ResNet and Wide ResNet can be
accurately represented by our encoding scheme, which is shown in Table
\ref{table:encoding_example}.

\subsection{NAS Search Algorithm}

\subsubsection{Non-dominated Sorting Algorithm}

Non-dominated sorting is mainly used to sort the solutions in population
according to the\textit{ Pareto dominance principle}, which plays
a very important role in the selection operation of many multi-objective
evolutionary algorithms. In non-dominated sorting, an individual A
is said to dominate another individual B, if and only if there is
no objective of A worse than that objective of B and there is at least
one objective of A better than that objective of B. Without loss of
generality, we assume that the solutions of a population $S$ can
be assigned to $K$ Pareto fronts $F_{i}$, $i=1,2,\ldots,K$. Non-dominated
sorting first selects all the non-dominated solutions from population
$S$ and assigns them to $F_{1}$ (the rank 1 front); it then selects
all the non-dominated solutions from the remaining solutions and assigns
them to $F_{2}$ (the rank 2 front); it repeats the above process
until all individuals have been assigned to a \textit{Pareto front}.

\subsubsection{NSGA-II: Elitist Non-Dominated Sorting Genetic Algorithm \label{subsec:NSGA-II:-Elitist-Non-Dominated}}

To solve the problem in Eq. \ref{eq:MOOP}, Elitist Non-Dominated
sorting genetic algorithm (NSGA-II) \cite{deb2000fast} is adopted
to optimize the \textit{Pareto front} $\mathcal{P}_{f}$ as shown
in Algorithm \ref{alg:Our-modified-NSGA-II}. In this paper, we choose
this kind of sample-based NAS algorithm instead of many popular parameter-sharing
NAS method. This is because we want to further analysis of the sampled
architectures and achieve insights and conclusions of the efficient
training. The main idea of NSGA-II is to rank the sampled architectures
by non-dominated sorting and preserve a group of elite architectures.
Then a group of new architectures is sampled and trained by mutation
of the current elite architectures on the $\mathcal{P}_{f}$ . The
algorithm can be paralleled on multiple computation nodes and lift
the $\mathcal{P}_{f}$ simultaneously. The mutation in the block-level
search space includes adding new skip-connection, modifying the current
operations and ratios. Meanwhile, the mutation in the macro-level
search space includes randomly adding or deleting one block in one
stage, exchanging the position of doubling channel block with its
neighbor, and modifying the base channels. This well-known NSGA-II
is easy to implement and we can easily monitor the improvement of
each iteration. The stop criterion depends on the time limit or the
computation cost constraints.

\begin{algorithm*}
\begin{centering}
\caption{\label{alg:Our-modified-NSGA-II}Our modified NSGA-II Searching Algorithm}
\par\end{centering}
\hspace*{\algorithmicindent} \textbf{Input} Stop criterion, Search Space, number of computation nodes N.
\begin{algorithmic}[1]
\State $t=0$
\State $P_{t}\leftarrow Random(A)$, generate a group of initial architectures.
\State Evaluate $P_{t}$
\While{not stop criterion}
\State Apply non-dominated sorting to $P_{t}$ to obtain non-dominated fronts $A_{i}^{*}$
\State Sort $A_{i}^{*}$ by Crowding distance and left top-N $A_{j}^{*}$ as Parents
\State Create new generation $Q_{t}$ by mutation on current $A_{j}^{*}$ 
\State Train the $Q_{t}$ on N computation nodes and Evaluate the accuracy of $Q_{t}$.
\State $P_{t+1}\leftarrow Q_{t} \cup P_{t}$
\State $t=t+1$
\EndWhile

\end{algorithmic}
\hspace*{\algorithmicindent} 
\textbf{Output} The Pareto Optimal Front $A_{i}^{*}$
\end{algorithm*}

\subsection{NAS Implementation Details\label{section:nas_search_process}}

\subsubsection{Block-level search}

In the phase of block-level search, a proxy task of ImageNet is created,
which is a subset sampled fromits training set. This subset constitutes
100 labels, each of which has 500 images as the training set, and
100 as the validation set. We call this dataset ImageNet-100 in the
following parts of this paper.

\begin{figure}
\begin{centering}
\vspace{-3mm}
\includegraphics[width=1\columnwidth]{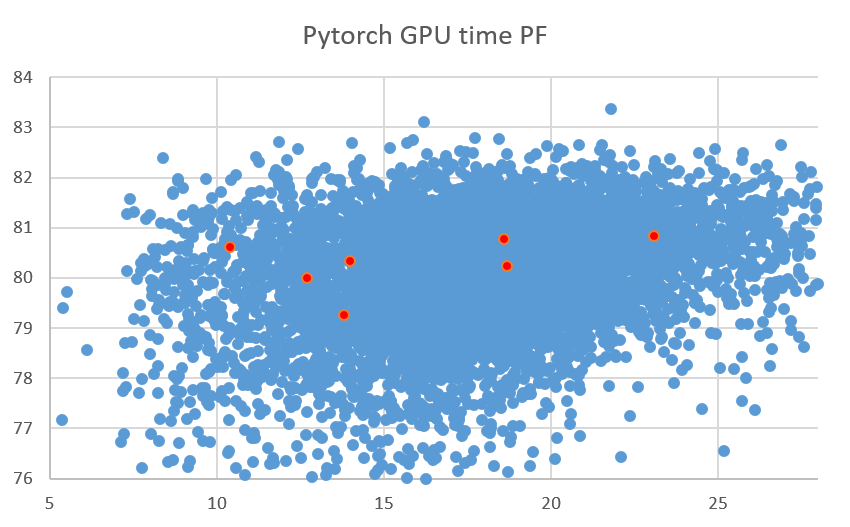} \vspace{1mm}
\par\end{centering}
\caption{\label{figure:micro_search_result}Results of the block-level search
in ImageNet-100. The y-axis denotes the accuracy and x-axis denotes
the latency. Blue dots are models searched in this step, while the
red ones are Basic Block with first channel 64, 128, 192; Inverted
Bottleneck Block (expansion rate 4) with first channel 64, 128; BottleNeck
Block (expansion rate 4) with first channel 256, 320. It can be found
that our algorithm can find more efficient block in the block-level
search.}

\centering{}\vspace{-2mm}
\end{figure}

To avoid interference with macro architecture, the macro-level architecture
is fixed to be the same as that of ResNet50. Each model is trained
by ImageNet-100 with a batch size of 32 for 90 epochs and learning
rate 0.1, which takes 3\textasciitilde 10 hours on a single NVIDIA
Tesla-V100 GPU. We do a random search at first, which uniformly samples
all the valid blocks in the search space. Evolutionary Algorithm (EA)
is then performed with three kinds of mutations: 1) replace one operator
with another; 2) change the output channel of one layer; 3) Add/remove/modify
a skip connection. We keep updating the Pareto Front between step
time and accuracy during the whole process. As a result, 10 blocks
are selected as the candidates for the following rounds. Practically,
during our search, the performance of early stop models aligns well
with the fully-train accuracy. We checked the Spearman Rank Correlation
for 103 architectures: $\rho=96.6\%$. Thus, using early stop can
greatly reduce the search cost by around 90\% while keeping our NAS
effective.

\subsubsection{Macro-level search}

We search the block-level architectures with the 10 blocks attained
by block-level search. Random search is adopted at first, where the
number of blocks is chosen randomly between 10 and 50, and the first
and last channel is drawn from $\{32,64,128\}$ and $\{512,1024,2048\}$
respectively. EA search is then applied, where the mutations allowed
are: 1) Add a `1'; 2) Remove a `1'; 3) Swap two different adjacent
numbers. Similar to block-level search, Pareto Front between step
time and accuracy is also kept updated.

Different from the previous phase, the whole ImageNet dataset is utilized
for training. Each model is trained with a batch size of 1024 and
a learning rate of 0.2 for 40 epochs. 

\subsection{ET-NAS: Model Zoo Information and their Encodings\label{subsec:ET-NAS:-Model-Zoo}}

\begin{table*}
\begin{centering}
\vspace{1mm}
{\footnotesize{}\label{table:nas_search_result}}{\scriptsize{}}%
\begin{tabular}{llrrrrrr}
\hline 
\multirow{2}{*}{{\scriptsize{}Model Name}} & \multirow{2}{*}{{\scriptsize{}Encoding}} & \multirow{2}{*}{{\scriptsize{}MParam}} & \multirow{2}{*}{{\scriptsize{}Gmac}} & \multirow{2}{*}{{\scriptsize{}MAct}} & \textbf{\scriptsize{}Top-1} & \textbf{\scriptsize{}inf time} & \textbf{\scriptsize{}Training step}\tabularnewline
 &  &  &  &  & \textbf{\scriptsize{}Acc} & \textbf{\scriptsize{}(ms)} & \textbf{\scriptsize{}time (ms)}\tabularnewline
\hline 
{\scriptsize{}ET-NAS-A} & {\scriptsize{}2-\_32\_2-11-112-1121112} & {\scriptsize{}2.6} & {\scriptsize{}0.23} & {\scriptsize{}1.3} & {\scriptsize{}62.06} & {\scriptsize{}5.30} & {\scriptsize{}14.74}\tabularnewline
{\scriptsize{}ET-NAS-B} & {\scriptsize{}031-\_32\_1-1-221-11121} & {\scriptsize{}3.9} & {\scriptsize{}0.39} & {\scriptsize{}1.3} & {\scriptsize{}66.92} & {\scriptsize{}5.92} & {\scriptsize{}15.78}\tabularnewline
{\scriptsize{}ET-NAS-C} & {\scriptsize{}011-\_32\_2-211-2-111122} & {\scriptsize{}7.1} & {\scriptsize{}0.58} & {\scriptsize{}2.0} & {\scriptsize{}71.29} & {\scriptsize{}8.94} & {\scriptsize{}26.28}\tabularnewline
{\scriptsize{}ET-NAS-D} & {\scriptsize{}031-\_64\_1-1-221-11121} & {\scriptsize{}15.2} & {\scriptsize{}1.55} & {\scriptsize{}2.6} & {\scriptsize{}74.46} & {\scriptsize{}14.54} & {\scriptsize{}36.30}\tabularnewline
{\scriptsize{}ET-NAS-E} & {\scriptsize{}011-\_64\_21-211-121-11111121} & {\scriptsize{}21.4} & {\scriptsize{}2.61} & {\scriptsize{}4.7} & {\scriptsize{}76.87} & {\scriptsize{}25.34} & {\scriptsize{}61.95}\tabularnewline
{\scriptsize{}ET-NAS-F} & {\scriptsize{}10001-\_64\_4-111-11122-1111111111111112} & {\scriptsize{}28.4} & {\scriptsize{}2.31} & {\scriptsize{}6.8} & {\scriptsize{}78.80} & {\scriptsize{}33.83} & {\scriptsize{}93.04}\tabularnewline
{\scriptsize{}ET-NAS-G} & {\scriptsize{}211-\_64\_41-211-121-11111121} & {\scriptsize{}49.3} & {\scriptsize{}5.68} & {\scriptsize{}8.4} & {\scriptsize{}80.41} & {\scriptsize{}53.08} & {\scriptsize{}133.97}\tabularnewline
{\scriptsize{}ET-NAS-H} & {\scriptsize{}10001-\_64\_4-111111111-211112111112-11111} & {\scriptsize{}44.0} & {\scriptsize{}5.33} & {\scriptsize{}10.9} & {\scriptsize{}80.92} & {\scriptsize{}76.80} & {\scriptsize{}193.40}\tabularnewline
{\scriptsize{}ET-NAS-I} & {\scriptsize{}02031-a02\_64\_111-2111-21111111111111111111111-211} & {\scriptsize{}72.4} & {\scriptsize{}13.13} & {\scriptsize{}14.6} & {\scriptsize{}81.38} & {\scriptsize{}94.60} & {\scriptsize{}265.13}\tabularnewline
{\scriptsize{}ET-NAS-J} & {\scriptsize{}211-\_64\_411-2111-21111111111111111111111-211} & {\scriptsize{}103.0} & {\scriptsize{}18.16} & {\scriptsize{}15.9} & {\scriptsize{}82.08} & {\scriptsize{}131.92} & {\scriptsize{}370.28}\tabularnewline
\multirow{2}{*}{{\scriptsize{}ET-NAS-K}} & {\scriptsize{}02031-a02\_64\_1121-111111111111111111111111111} & \multirow{2}{*}{{\scriptsize{}87.3}} & \multirow{2}{*}{{\scriptsize{}27.51}} & \multirow{2}{*}{{\scriptsize{}31.3}} & \multirow{2}{*}{{\scriptsize{}82.42}} & \multirow{2}{*}{{\scriptsize{}185.75}} & \multirow{2}{*}{{\scriptsize{}505.00}}\tabularnewline
 & {\scriptsize{}-21111111211111-1} &  &  &  &  &  & \tabularnewline
\multirow{2}{*}{{\scriptsize{}ET-NAS-L}} & {\scriptsize{}23311-a02c12\_64\_211-2111} & \multirow{2}{*}{{\scriptsize{}130.4}} & \multirow{2}{*}{{\scriptsize{}23.46}} & \multirow{2}{*}{{\scriptsize{}19.4}} & \multirow{2}{*}{\textbf{\scriptsize{}82.65}} & \multirow{2}{*}{{\scriptsize{}191.89}} & \multirow{2}{*}{{\scriptsize{}542.52}}\tabularnewline
 & {\scriptsize{}-21111111111111111111111-211} &  &  &  &  &  & \tabularnewline
\hline 
\end{tabular}{\scriptsize\par}
\par\end{centering}
\begin{centering}
\vspace{1mm}
\par\end{centering}
\centering{}{\footnotesize{}\caption{The searched optimal efficient training models ''ET-NAS'' found by
our NAS search. `Acc' means the accuracy evaluated on the ImageNet;
inference time and step time are measured in ms on single Nvidia V100,
with a batch size of 64. By observing the optimal model, smaller models
should use simpler blocks while bigger models prefer complex blocks.}
}{\footnotesize\par}
\end{table*}

After the NAS process done in Section \ref{section:nas_search_process},
12 models are selected as our model zoo ET-NAS of fine-tuning. Details
of these models are shown in Table \ref{table:nas_search_result}.
The inference time and step time are measured in ms on a single Nvidia
V100, with batch size of 64. The resolution follows the standard setting
of ImageNet: 224x224. By observing the optimal model in the table,
smaller models should use simpler blocks while bigger models prefer
complex blocks.

\begin{figure*}[t]
\begin{centering}
\vspace{-2mm}
\par\end{centering}
\begin{centering}
\includegraphics[width=0.9\columnwidth]{compare_with_nas_training}\includegraphics[width=0.9\columnwidth]{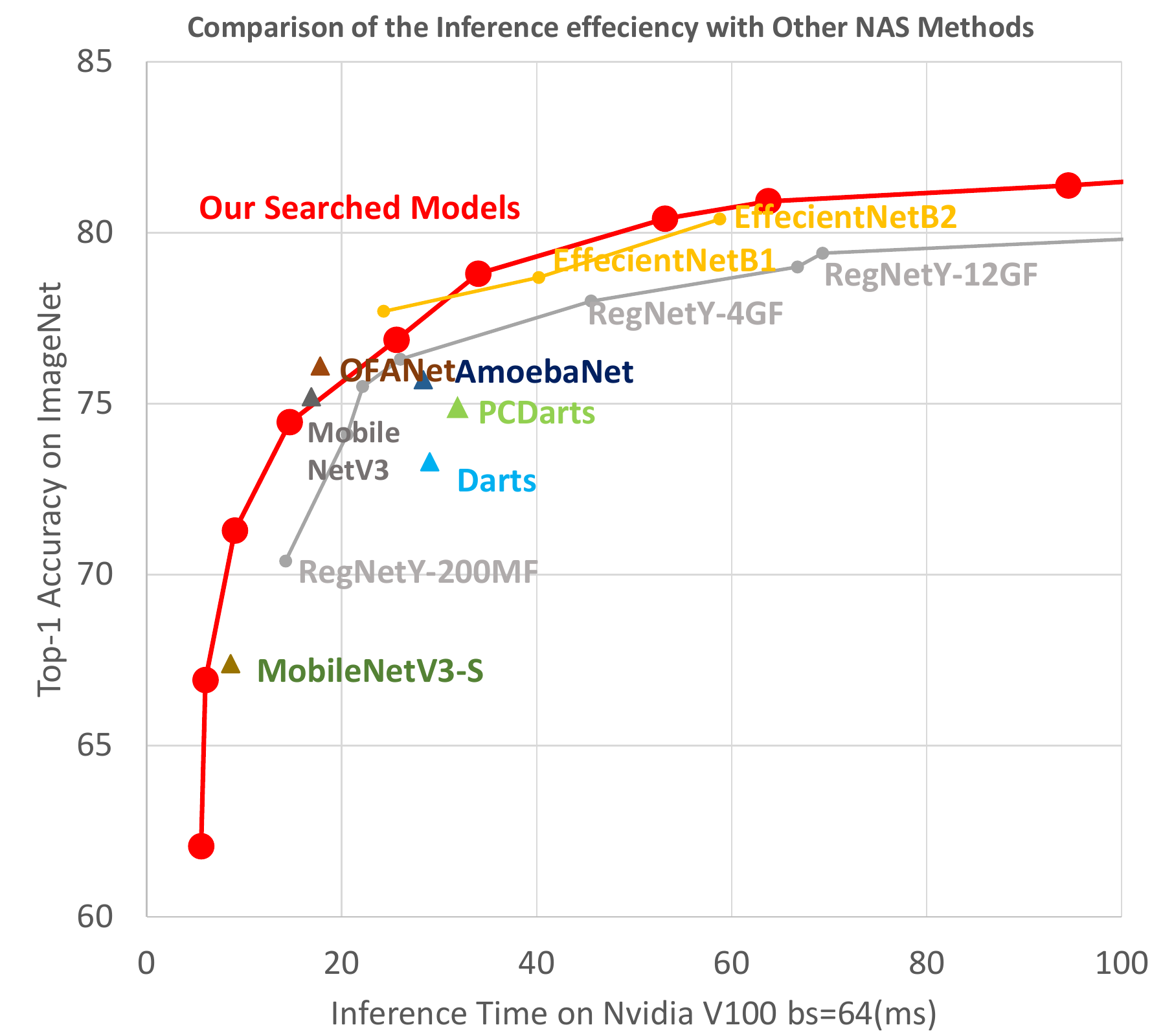}
\par\end{centering}
\begin{centering}
\vspace{1mm}
\par\end{centering}
\caption{{\small{}\label{fig:Comparison-of-the-efficiency-1}Comparison of
the training and inference efficiency of our searched models (ET-NAS)
with SOTA NAS models on ImageNet. We further compared our models with
8 other NAS results. It can be found that our method is more training
efficient than some recent evolution-based NAS methods such as AmoebaNet
\cite{real2019regularized}, OFANet \cite{cai2019once} because of
our effective search space.}}

\centering{}\vspace{-1mm}
\end{figure*}

Figure \ref{fig:Comparison-of-the-efficiency-1} shows the comparison
of our ET-NAS models with other SOTA ImageNet models. Inference time
and training step time are measured in ms on a single Nvidia V100,
with $bs=64$. Our ET-NAS series show superior performance comparing
to RegNet, EfficientNet series. Comparing to some EA-based NAS methods
such as OFANet and Amoebanet, our method is also efficient in terms
of training. We found that there exists a performance ranking gap
between inference time and training step time in Figure \ref{fig:Comparison-of-the-efficiency-1}.
This is mainly due to the depth and the main type of operation of
the models. We found that deeper networks with separable conv such
as EfficientNet/MobileNet have a larger training-step-time/inference-time
ratio comparing to our models (shallower\&more common conv).

\subsubsection{What makes a network efficient-training?\label{subsec:Analysis:-What-kind}}

To answer this question, we first need to define a score for the efficiency
of the searched models $\mathcal{A}$. In MOOP, the goodness of a
solution is determined by \textit{dominance}. Thus, we can use the
non-dominated sorting algorithm to sort the $\mathcal{A}$ according
to the Pareto dominance principle. Each architecture is assigned to
one \textit{Pareto front} and the rank $R_{\mathcal{P}}$ of that
\textit{Pareto front }can be regarded as the goodness of a solution,
in our case, the efficiency. We then defined the \textit{efficiency
score} of $\mathcal{A}$ as: $s_{E}(\mathcal{A})=-\frac{R_{\mathcal{P}}(\mathcal{A})-\mathrm{mean}(R_{\mathcal{P}}(\mathcal{A}))}{\mathrm{std}(R_{\mathcal{P}}(\mathcal{A}))}.$
Since \textit{Pareto optimal front} is the Rank 1 \textit{Pareto front},
larger efficiency score $s_{E}(\mathcal{A})$ means better efficiency.

\begin{table*}
\begin{centering}
\vspace{-1mm}
\par\end{centering}
\begin{centering}
\renewcommand\arraystretch{1}\tabcolsep 0.04in%
\begin{tabular}{ccccc|ccccc}
\hline 
\multicolumn{3}{c}{{\scriptsize{}Block-level Regression Analysis}} & {\scriptsize{}n=5500} & {\scriptsize{}R-sq=56\%} & \multicolumn{3}{c}{{\scriptsize{}Macro-level Regression Analysis}} & {\scriptsize{}n=1200} & {\scriptsize{}R-sq=71\%}\tabularnewline
\hline 
\textbf{\scriptsize{}Terms} & \textbf{\scriptsize{}Coef} & \textbf{\scriptsize{}SE Coef} & \textbf{\scriptsize{}T-Value} & \textbf{\scriptsize{}P-Value} & \textbf{\scriptsize{}Terms} & \textbf{\scriptsize{}Coef} & \textbf{\scriptsize{}SE Coef} & \textbf{\scriptsize{}T-Value} & \textbf{\scriptsize{}P-Value}\tabularnewline
\hline 
\textit{\scriptsize{}OP1 Channel Change Ratio} & {\scriptsize{}-0.183} & {\scriptsize{}0.010} & {\scriptsize{}-28.57} & {\scriptsize{}0.06} & \textbf{\textit{\scriptsize{}Channel Size}} & \textbf{\scriptsize{}-0.210} & {\scriptsize{}0.023} & {\scriptsize{}-9.26} & {\scriptsize{}0.00}\tabularnewline
\textit{\scriptsize{}OP2 Channel Change Ratio} & {\scriptsize{}-0.168} & {\scriptsize{}0.010} & {\scriptsize{}-27.75} & {\scriptsize{}0.02} & \textbf{\textit{\scriptsize{}Double Channel Position 1}} & \textbf{\scriptsize{}-0.110} & {\scriptsize{}0.026} & {\scriptsize{}-4.17} & {\scriptsize{}0.00}\tabularnewline
\textbf{\textit{\scriptsize{}Num of skip connection (add)}} & \textbf{\scriptsize{}-0.272} & {\scriptsize{}0.018} & {\scriptsize{}-15.12} & {\scriptsize{}0.00} & \textit{\scriptsize{}Double Channel Position 2} & {\scriptsize{}0.035} & {\scriptsize{}0.030} & {\scriptsize{}1.15} & {\scriptsize{}0.25}\tabularnewline
\textbf{\textit{\scriptsize{}Num of skip connection (concat)}} & \textbf{\scriptsize{}-0.362} & {\scriptsize{}0.018} & {\scriptsize{}-19.93} & {\scriptsize{}0.00} & \textit{\scriptsize{}Double Channel Position 3} & {\scriptsize{}-0.016} & {\scriptsize{}0.030} & {\scriptsize{}-0.54} & {\scriptsize{}0.59}\tabularnewline
\textit{\scriptsize{}Output\_channel} & {\scriptsize{}-0.539} & {\scriptsize{}0.010} & {\scriptsize{}-53.57} & {\scriptsize{}0.00} & \textbf{\textit{\scriptsize{}Double Channel Position 4}} & \textbf{\scriptsize{}0.224} & {\scriptsize{}0.025} & {\scriptsize{}9.14} & {\scriptsize{}0.00}\tabularnewline
\textit{\scriptsize{}conv3x3 (ref)} & {\scriptsize{}0.000} & {\scriptsize{}-} & {\scriptsize{}-} & {\scriptsize{}-} & \textit{\scriptsize{}Double Channel Position 5} & {\scriptsize{}0.036} & {\scriptsize{}0.022} & {\scriptsize{}1.62} & {\scriptsize{}0.11}\tabularnewline
\textit{\scriptsize{}conv1x1} & {\scriptsize{}-0.037} & {\scriptsize{}0.033} & {\scriptsize{}-0.72} & {\scriptsize{}0.08} & \textbf{\textit{\scriptsize{}Num block in Stage-1}} & \textbf{\scriptsize{}-0.562} & {\scriptsize{}0.023} & {\scriptsize{}-24.69} & {\scriptsize{}0.00}\tabularnewline
\textit{\scriptsize{}conv3x3, w group=2} & {\scriptsize{}0.190} & {\scriptsize{}0.035} & {\scriptsize{}5.49} & {\scriptsize{}0.00} & \textbf{\textit{\scriptsize{}Num block in Stage-2}} & \textbf{\scriptsize{}-0.139} & {\scriptsize{}0.023} & {\scriptsize{}-6.04} & {\scriptsize{}0.00}\tabularnewline
\textbf{\textit{\scriptsize{}conv3x3, w group=4}} & \textbf{\scriptsize{}0.295} & {\scriptsize{}0.034} & {\scriptsize{}8.77} & {\scriptsize{}0.00} & \textbf{\textit{\scriptsize{}Num block in Stage-3}} & \textbf{\scriptsize{}0.044} & {\scriptsize{}0.024} & {\scriptsize{}1.88} & {\scriptsize{}0.06}\tabularnewline
\textbf{\textit{\scriptsize{}Separable conv3x3}} & \textbf{\scriptsize{}-0.200} & {\scriptsize{}0.034} & {\scriptsize{}-5.91} & {\scriptsize{}0.00} & \textit{\scriptsize{}Num block in Stage-4} & {\scriptsize{}-0.010} & {\scriptsize{}0.024} & {\scriptsize{}-0.42} & {\scriptsize{}0.67}\tabularnewline
\hline 
\end{tabular}
\par\end{centering}
\begin{centering}
\vspace{1mm}
\par\end{centering}
\centering{}\caption{\label{tab:Regression-Analysis}Regression Analysis: what makes a
network efficient-training? We exam the effect of each component of
network on the efficiency score. ``Coef'' and ``SE Coef'' are
the estimated regression coefficient and standard error. ``T-Value''/``P-Value''
shows the significance of the variables.}
\end{table*}

Then we perform a multivariate linear regression analysis on the $\mathcal{A}_{S}$.
According to our search space, ordinal/nominal variables that describe
the model are denoted as predictors to fit the $s_{E}(\mathcal{A})$.
Table \ref{tab:Regression-Analysis} shows the coefficients from the
regression analysis on both block-level and macro-level designs. Positive
coefficients indicate a positive relationship. \textquotedblleft P-Value\textquotedblright{}
shows the significance of the variables. We summarize and highlight
several noteworthy conclusions uncovered by our analysis:

\begin{itemize}[noitemsep, nolistsep, leftmargin=*]

\item By observing optimal $\mathcal{A}_{i}^{*}$, smaller models
should use simpler blocks while bigger models prefer complex blocks.
Simply increasing depth/width to expand the model in \cite{Tan2019}
may not be optimal.

\item Adding additional skip connections will decrease the training
efficiency of the model (The Coef is significantly negative). Using
``add'' to combine the features is more efficient than ``concat''.

\item Using ``conv3x3, w group=4'' is the best operation among
the searched operations (Coef is 0.295). Separable conv3x3 is not
efficient for training (Coef is -0.2).

\item The first double-channel position should be more close to the
beginning of the network, while the final double channel-position
should be delayed to the end of the network.

\item Fewer blocks should be assigned to the first two stages. More
should be assigned to the 3rd stage.

\end{itemize}

\begin{algorithm*}
\caption{\label{alg:Efficient-Training-Model}Efficient Training Model Zoo
(ET-NAS) Creation}

	\begin{algorithmic}[1]
		\Require Block/Macro Search Space $\mathcal{S}_i, \mathcal{S}_a$, Stop Criterion $\Gamma$, $\#$Computation Nodes $K$, Sensitive Factor $\epsilon$, $\#$Block Architectures $M$, $\#$Models in Model Zoo $N$.
		\Ensure Final Model Zoo $Z_{oo}$
		\Procedure{BlockSearch}{$\mathcal{S}_i, \Gamma, K, \epsilon, M$}
		\State $P_f \gets$ \Call{NSGA-II}{$\Gamma$, $\mathcal{S}_i, K, \epsilon$} \Comment{Our modified NSGA-II, see Algorithm \ref{alg:Our-modified-NSGA-II}}
		\State $Cells \gets$ \Call{MostCommon}{$P_f, M$} \Comment{Most common $M$ cells from $P_f$}
		\EndProcedure
		\Procedure{MacroSearch}{$\mathcal{S}_a, \Gamma, K, \epsilon, N$}
		\State $P_f \gets$ \Call{NSGA-II}{$\Gamma$, $\mathcal{S}_a(Cells), K, \epsilon$}
		\State $Z_{oo} \gets$ \Call{NsgaSort}{$P_f, N, \epsilon$} \Comment{Choose models based on crowding-distance}
		\EndProcedure
	\end{algorithmic}
\end{algorithm*}

\begin{algorithm*}
\caption{\label{alg:Online-Fine-Tuning-schedule}Online Fine-Tuning schedule
Generator Training, Prediction, and Update}

	\begin{algorithmic}[1]
		\Require Model Zoo $Z_{oo}$, Time Evaluator $T_{s}$, Acc Evaluator $T_{r}$, Hyper-parameter Space $\mathcal{S}_{HP}$, Known Datasets $D_{old}$, New Dataset $D_{new}$, $\#$Meta-data $H_M$, Time Constraint $T_l$, $\#$Configurations $H$.
		\Ensure Optimal Model $\mathcal{A}^*$, Hyper-parameters $Regime_{FT}^*$, Predictor $Acc_P$
		\Procedure{OfflineTraining}{$Z_{oo}, T_{r}, \mathcal{S}_{HP}, D_{old}, H_M$}
		\State $MetaData \gets \emptyset, Acc_P \gets$ \Call{AdaptiveMLP}{.} \Comment{Initialize default predictor}
		\For{$D \in D_{old}, i \gets 1$ to $H_M$}
			\State $\mathcal{A}, Regime_{FT} \gets$ \Call{Random}{$Z_{oo}, S_{HP}$} \Comment{Randomly select from search space}
			\State $Acc \gets$ \Call{$T_r$}{$D, \mathcal{A}, Regime_{FT}$} \Comment{Train with selected configuration}
			\State $MetaData \gets MetaData \cup \{(\mathcal{A}, Regime_{FT}, Acc)\}$ \Comment{Add this result to meta-data}
		\EndFor
		\State $Acc_P \gets$ \Call{Train}{$Acc_P, MetaData$} \Comment{Train predictor with all meta-data}
		\EndProcedure
		
		\Procedure{OnlinePrediction}{$Z_{oo}, D_{new}, Acc_P, T_l, T_s, H$}
		\State $MetaData \gets \emptyset$
		\For{$i \gets 1$ to $H$}
			\State $\mathcal{A} \gets$ \Call{Random}{$Z_{oo}$}
			\State $Epoch \gets T_l \div$ \Call{$T_{s}$}{$\mathcal{A}, D_{new}$} \Comment{Always choose the largest epoch within $T_l$}
			\State $Regime_{FT} \gets$ \Call{Random}{$S_{HP}|Epoch$} \Comment{Randomly select condition on $Epoch$}
			\State $MetaData \gets MetaData \cup \{(\mathcal{A}, Regime_{FT})\}$
		\EndFor
		\State $\mathcal{A}^*, Regime_{FT}^* \gets$ \Call{Predict}{$G, MetaData$} \Comment{Choose the optimal from $H$ configs}
		\State $Acc \gets$ \Call{$T_r$}{$D_{new}, \mathcal{A}^*, Regime_{FT}^*$} 
		\State $Acc_P \gets$ \Call{Train}{$Acc_P, \{\mathcal{A}^*, Regime_{FT}^*, Acc\}$} \Comment{Improve predictor with this meta-data}
		\EndProcedure
	\end{algorithmic}
\end{algorithm*}

\subsection{CO$_{2}$ consumption Analysis}

Fine-tuning from the pretrained ImageNet/language model is a de-facto
practice in the deep learning field (CV/NLP). Our NASOA improves the
efficiency of fine-tuning which has the potentials to greatly reduce
computational cost in GPU clusters/cloud computing. According to a
recent study \cite{strubell2019energy}, developing and tuning for
one typical R\&D project \cite{strubell2018linguistically} in Google
Cloud computing needs about \$250k cost, 82k kWh electricity, and
123k lbs CO$_{2}$ emission, which equals to the CO$_{2}$ consumption
of air traveling (NY$\leftrightarrow$SF) 62 times. Among most of
them, 123 hyperparameter grid searches were performed for new datasets,
resulting in 4789 jobs in total. It is believed that the proposed
faster fine-tuning pipeline can save up to 40x computational cost
among them. Furthermore, we have released all the searched efficient
models to help the public skipping the computation-heavy NAS stage
and directly enjoy the benefit of our methods. In conclusion, our
NASOA is meaningful for environment protection and energy saving.

\subsection{Detailed Algorithms of NASOA\label{subsec:Detailed-Algorithms-of}}

Detailed algorithms of our Model Zoo (ET-NAS) search can be found
in Algorithm \ref{alg:Efficient-Training-Model}. The pseudo code
of online fine-tuning schedule generator training, prediction, and
update can be found in Algorithm \ref{alg:Online-Fine-Tuning-schedule}.

\section{Proof of the Theoretical Results}

Let the $\alpha$ and $\mathcal{L}$ denote as
\begin{align*}
\alpha & =\left(\alpha_{1},\alpha_{2},\ldots,\alpha_{L}\right)^{\mathrm{T}},\\
\mathcal{L} & =\left(\mathcal{L}_{1},\mathcal{L}_{2},\ldots,\mathcal{L}_{L}\right)^{\mathrm{T}}
\end{align*}
 where $\mathcal{L}_{l}=\mathcal{L}(f_{l},Acc_{gt})$ for $l=1,2,\ldots,L.$
At time $0\leq t\leq T$, we denote $\alpha$ and $\mathcal{L}$ as
$\alpha^{(t)}$ and $\mathcal{L}^{(t)}$ respectively.
\begin{thm*}
Suppose the number of layers $L$ is a fixed integer, the training
time $T$ is sufficiently large and the loss function $\mathcal{L}(f_{l},Acc_{gt})$
is bounded in $[0,1]$. The sequence of the weight vectors: 
\[
\{\alpha^{(1)},\alpha^{(1)},\ldots,\alpha^{(T)}\},
\]
is learned by the Hedge Backpropagation. The initialized weight vector
$\alpha^{(1)}$ is the uniform discrete distribution 
\[
\alpha^{(1)}=(\frac{1}{L},\frac{1}{L},\ldots,\frac{1}{L}).
\]
The discount rate $\beta$ is fixed during the training procedure
and is taken to be $\sqrt{T}/(\sqrt{T}+C)$ given $T$, where $C$
is a fixed constant. Then the average regret of the online learning
algorithm for modelling $Acc(\mathcal{A}_{i}^{FT},D_{val})$ satisfies
\[
\frac{1}{T}\sum_{t=1}^{T}\left(\alpha^{(t)}\right)^{\mathrm{T}}\mathcal{L}^{(t)}-\min_{\alpha}\frac{1}{T}\sum_{t=1}^{T}\alpha^{\mathrm{T}}\mathcal{L}^{(t)}\leq O(\frac{1}{\sqrt{T}}).
\]
\end{thm*}

{\bf Proof:} This theorem is derived from Theorem~1 of \cite{freund1999adaptive} and Theorem~2 of \cite{freund1997decision}. Write the KullbackLeibler divergence between two weight vectors $\alpha$ and $\alpha'$ as
\benn
KL(\alpha \| \alpha') = \sum_{l=1}^L \alpha_l \ln (\frac{\alpha_l}{\alpha'_l}).
\eenn 
For any weight vector $\alpha$, we have
\benrr
&& KL(\alpha\|\alpha^{(t+1)}) - KL(\alpha\|\alpha^{(t)}) \\
&=& \sum_{l=1}^L \alpha_l \ln (\frac{\alpha_l}{\alpha^{(t+1)}_l}) - \sum_{l=1}^L \alpha_l \ln (\frac{\alpha_l}{\alpha^{(t)}_l}) \\
&=& \sum_{l=1}^L \alpha_l \ln (\frac{\alpha^{(t)}_l}{\alpha^{(t+1)}_l}) \\
&=& \sum_{l=1}^L \alpha_l \ln \big(\frac{ \sum_{l'=1^L} \alpha_{l'}^{(t)} \beta^{\mathcal{L}^{(t)}_{l'}} }{ \beta^{\mathcal{L}^{(t)}_l}}\big) \\
&=& \sum_{l=1}^L \alpha_l \ln (\sum_{l'=1^L} \alpha_{l'}^{(t)} \beta^{\mathcal{L}^{(t)}_{l'}}) - \ln (\beta) \sum_{l=1}^L \alpha_l \mathcal{L}^{(t)}_{l} \\
&=& \ln (\sum_{l'=1^L} \alpha_{l'}^{(t)} \beta^{\mathcal{L}^{(t)}_{l'}}) - \ln (\beta) \alpha^\T \mathcal{L}^{(t)}.
\eenrr
By convexity, 
$
\beta^{\mathcal{L}^{(t)}_{l'}} \leq 1 - (1-\beta) \mathcal{L}^{(t)}_{l'}.
$
Thus,
\benrr
&& \ln (\sum_{l'=1^L} \alpha_{l'}^{(t)} \beta^{\mathcal{L}^{(t)}_{l'}}) \\
&\leq& \ln \big(\sum_{l'=1^L} \alpha_{l'}^{(t)} (1 - (1-\beta) \mathcal{L}^{(t)}_{l'}) \big) \\
&=& \ln \big(1- (1-\beta)(\alpha^{(t)})^\T \mathcal{L}^{(t)}\big).
\eenrr
Then we have
\benrr
&&KL(\alpha\|\alpha^{(t+1)}) - KL(\alpha\|\alpha^{(t)}) \\
&\leq& \ln \big(1 - (1-\beta)(\alpha^{(t)})^\T \mathcal{L}^{(t)}\big) -  \ln (\beta) \alpha^\T \mathcal{L}^{(t)} \\
&\leq& - (1-\beta)(\alpha^{(t)})^\T \mathcal{L}^{(t)} -  \ln (\beta) \alpha^\T \mathcal{L}^{(t)}, 
\eenrr
where the second inequality holds since $\ln (1-x) \leq -x$ for $x<1$ and $0 \leq (1-\beta)(\alpha^{(t)})^\T \mathcal{L}^{(t)} \leq 1.$
Taking summation from $t=1$ to $T$ for both sides of the inequality,
\benrr
&& KL(\alpha\|\alpha^{(T+1)}) - KL(\alpha\|\alpha^{(1)}) \\
&\leq& - (1-\beta) \sum_{t=1}^T (\alpha^{(t)})^\T \mathcal{L}^{(t)} -  \ln (\beta) \sum_{t=1}^T \alpha^\T \mathcal{L}^{(t)}.
\eenrr
By rearranging the inequality, we have 
\benn
\frac{1}{T} \sum_{t=1}^T (\alpha^{(t)})^\T \mathcal{L}^{(t)} \leq I_1 + I_2 + I_3,
\eenn
where
\benrr
I_1 &=& -\frac{\ln(\beta)}{(1-\beta)} \frac{1}{T} \sum_{t=1}^T \alpha^\T \mathcal{L}^{(t)} \\
I_2 &=& \frac{1}{(1-\beta)} \frac{1}{T} KL(\alpha\|\alpha^{(1)}) \\
I_3 &=& - \frac{1}{(1-\beta)T} KL(\alpha\|\alpha^{(T+1)}).
\eenrr
First, $I_3$ is non-positive. Thus we focus on $I_1$ and $I_2$.
Let $\beta = 1/(1+c_t).$  Then the term $I_1$ is bounded above by
\benrr
I_1 &=& -\frac{\ln(\beta)}{(1-\beta)} \frac{1}{T} \sum_{t=1}^T \alpha^\T \mathcal{L}^{(t)} \\
&\leq& \frac{(1 + \beta)}{2 \beta} \frac{1}{T} \sum_{t=1}^T \alpha^\T \mathcal{L}^{(t)} \\
&=& \big(1 + \frac{c_t}{2}\big) \frac{1}{T} \sum_{t=1}^T \alpha^\T \mathcal{L}^{(t)}
\eenrr
where the inequality holds since $-\ln (\beta) \leq (1-\beta^2)/(2\beta).$
Note that the loss function $\mathcal{L}$ is bounded. So
\benrr
I_1 =  \frac{1}{T} \sum_{t=1}^T \alpha^\T \mathcal{L}^{(t)} + O(c_t)
\eenrr
On the other hand,
\benrr
I_2 &=& \frac{1}{(1-\beta)T}KL(\alpha\|\alpha^{(1)}) \\
&=& (\frac{1}{T} + \frac{1}{c_t T}) KL(\alpha\|\alpha^{(1)}) \\
& = & O(\frac{1}{T}) + O(\frac{1}{c_t T}),
\eenrr
where the third equality hold since $\alpha^{(1)}$ is the uniform distribution and $KL(\alpha\|\alpha^{(1)})\leq \ln L$ for any $\alpha.$
Trading off the convergence rate of $O(c_t)$ and $O((c_t T)^{-1})$, the optimal rate of $c_t$ is $O(1/\sqrt T).$ When $T$ is sufficiently large, 
\benn
I_1 = \frac{1}{T}  \sum_{t=1}^T \alpha^\T \mathcal{L}^{(t)} + O(\frac{1}{\sqrt T}),
\eenn
and 
\benn
I_2 = O(\frac{1}{T}) + O(\frac{1}{\sqrt T}).
\eenn
Combining the results of $I_1$, $I_2$ and $I_3$, we have
\benrr
\frac{1}{T} \sum_{t=1}^T (\alpha^{(t)})^\T \mathcal{L}^{(t)} \leq \frac{1}{T}  \sum_{t=1}^T \alpha^\T \mathcal{L}^{(t)} + O(\frac{1}{\sqrt T})
\eenrr
for any weight vector $\alpha.$
Hence
\benrr
\frac{1}{T} \sum_{t=1}^T (\alpha^{(t)})^\T \mathcal{L}^{(t)} \leq \min_{\alpha}\frac{1}{T}  \sum_{t=1}^T \alpha^\T \mathcal{L}^{(t)} + O(\frac{1}{\sqrt T}).
\eenrr

\section{Implement Details of HPO methods}

We use BOHB and random search in our experiments as the HPO baseline.
As stated in Section 4.4, The shortest/longest time constraint (budget)
is defined as the time of fine-tuning 10/50 epochs for ResNet18/ResNet101
and the rest are equally divided into the log-space, which can be
represented as: $t_{x}=t_{0}*(t_{3}/t_{0})^{x/t_{3}}$, where $x=[0,1,2,3]$,
$t_{0}$is the time to train ResNet18 for 10 epochs and $t_{3}$ is
the time to train ResNet101 for 50 epochs.

We only compare the HPO setting under the same max computational budgets
equal to $t_{1}$ in Table 5 (left). For random search, we randomly
sample candidates from predefined search space until reaching the
max computational budget. And for BOHB, we use the opensource implementation
of BOHB at \url{https://github.com/automl/HpBandSter}. We set random
fraction=0.3, percent of good observations=15\%, min budget=25\% and
max budget=100\% with respect to our max computational budget. And
we use BOHB with 40x computational cost than our proposed methods
with different model zoos. The results are presented in main text.
\end{document}

%% file: math.tex
\usepackage{amsmath,amsfonts,bm}

\def\eqref#1{equation~\ref{#1}}

\def\1{\bm{1}}

\DeclareMathAlphabet{\mathsfit}{\encodingdefault}{\sfdefault}{m}{sl}
\SetMathAlphabet{\mathsfit}{bold}{\encodingdefault}{\sfdefault}{bx}{n}

\newcommand{\T}{{\mathrm{\scriptscriptstyle T}}}

\newcommand{\benr}{\begin{eqnarray}}
\newcommand{\eenr}{\end{eqnarray}}
\newcommand{\benrr}{\begin{eqnarray*}}
\newcommand{\eenrr}{\end{eqnarray*}}
\newcommand{\ben}{\begin{equation}}
\newcommand{\een}{\end{equation}}
\newcommand{\benn}{\begin{equation*}}
\newcommand{\eenn}{\end{equation*}}